\documentclass[preprint,12pt]{elsarticle}
\pdfoutput=1

\usepackage{amssymb}
\usepackage{amsmath}

\usepackage{booktabs}
\usepackage{array}
\usepackage{multirow}
\usepackage{natbib}
\usepackage{hyperref}
\usepackage{xcolor}         
\usepackage[colorinlistoftodos,textsize=scriptsize]{todonotes}

\newcolumntype{x}[1]{>{\centering\arraybackslash\hspace{0pt}}p{#1}}

\journal{journal}

\begin{document}

\begin{frontmatter}



\title{Unlocking large-scale crop field delineation in smallholder farming systems with transfer learning and weak supervision}


\author[inst1,inst2,inst3]{Sherrie Wang}
\author[inst4]{Fran\c{c}ois Waldner}
\author[inst3]{David B. Lobell}

\affiliation[inst1]{organization={Goldman School of Public Policy, UC Berkeley},
            addressline={2607 Hearst Ave}, 
            city={Berkeley},
            postcode={94720}, 
            state={CA},
            country={USA}}
\affiliation[inst2]{organization={Institute for Computational and Mathematical Engineering, Stanford University},
            addressline={475 Via Ortega}, 
            city={Stanford},
            postcode={94305}, 
            state={CA},
            country={USA}}
\affiliation[inst3]{organization={Department of Earth System Science, Stanford University},
            addressline={473 Via Ortega}, 
            city={Stanford},
            postcode={94305}, 
            state={CA},
            country={USA}}
\affiliation[inst4]{organization={European Commission Joint Research Centre},
            addressline={Via Enrico Fermi 2749}, 
            city={Ispra},
            postcode={21027}, 
            state={VA},
            country={Italy}}

\begin{abstract}
Crop field boundaries aid in mapping crop types, predicting yields, and delivering field-scale analytics to farmers. Recent years have seen the successful application of deep learning to delineating field boundaries in industrial agricultural systems, but field boundary datasets remain missing in smallholder systems due to (1) small fields that require high resolution satellite imagery to delineate and (2) a lack of ground labels for model training and validation. In this work, we combine transfer learning and weak supervision to overcome these challenges, and we demonstrate the methods' success in India where we efficiently generated 10,000 new field labels. Our best model uses 1.5m resolution Airbus SPOT imagery as input, pre-trains a state-of-the-art neural network on France field boundaries, and fine-tunes on India labels to achieve a median Intersection over Union (IoU) of 0.86 in India. If using 4.8m resolution PlanetScope imagery instead, the best model achieves a median IoU of 0.72. Experiments also show that pre-training in France reduces the number of India field labels needed to achieve a given performance level by as much as $20\times$ when datasets are small. These findings suggest our method is a scalable approach for delineating crop fields in regions of the world that currently lack field boundary datasets. 
We publicly release the 10,000 labels and delineation model to facilitate the creation of field boundary maps and new methods by the community.
\end{abstract}


\end{frontmatter}


\section{Introduction}
\label{sec:intro}

A crop field is the basic unit of management in agriculture. Delineating field boundaries allows one to capture the size, shape, and spatial distribution of agricultural fields, which are important characteristics of rural landscapes \cite{fritz2015mapping, yan2016conterminous}. By enabling field-level analysis, field boundaries are helpful inputs to crop type mapping \cite{dewit2004efficiency, cai2018high, wang2019crop, lambert2018estimating}, yield mapping \cite{lobell2015scalable, maestrini2018predicting, kang2019field, donohue2018towards, schwalbert2020satellite, algaadi2016prediction}, and digital agriculture services \cite{bramley2019farmer}. Previous work has also related field size to productivity \cite{carter1984identification, chand2011farm, rada2019new}, pest and disease spread \cite{segoli2012should, schaafsma2005effect}, and species diversity \cite{fahrig2015farmlands, salek2018bringing}.

Despite their usefulness, field boundary datasets remain unavailable in most countries. 
Where they do exist, field boundaries are either gathered through ground surveys \cite{persello2019delineation}, submitted by farmers to a statistical agency \cite{france, donohue2018towards}, or extracted from very high resolution remote sensing imagery \cite{waldner2020deep, vlachopoulos2020delineation, north2019boundary}. All of these methods are expensive, labor-intensive, and require sophisticated statistical infrastructure; as a result, datasets are most likely to exist in high-income countries. In many medium- and low-income countries, the capacity to conduct surveys or fly imaging aircraft is limited or nonexistent. To our knowledge, there is no large-scale dataset in any smallholder region in the Global South, despite the importance of agricultural management and productivity in these regions. 

\begin{figure}
	\centering
	\includegraphics[width=0.9\linewidth]{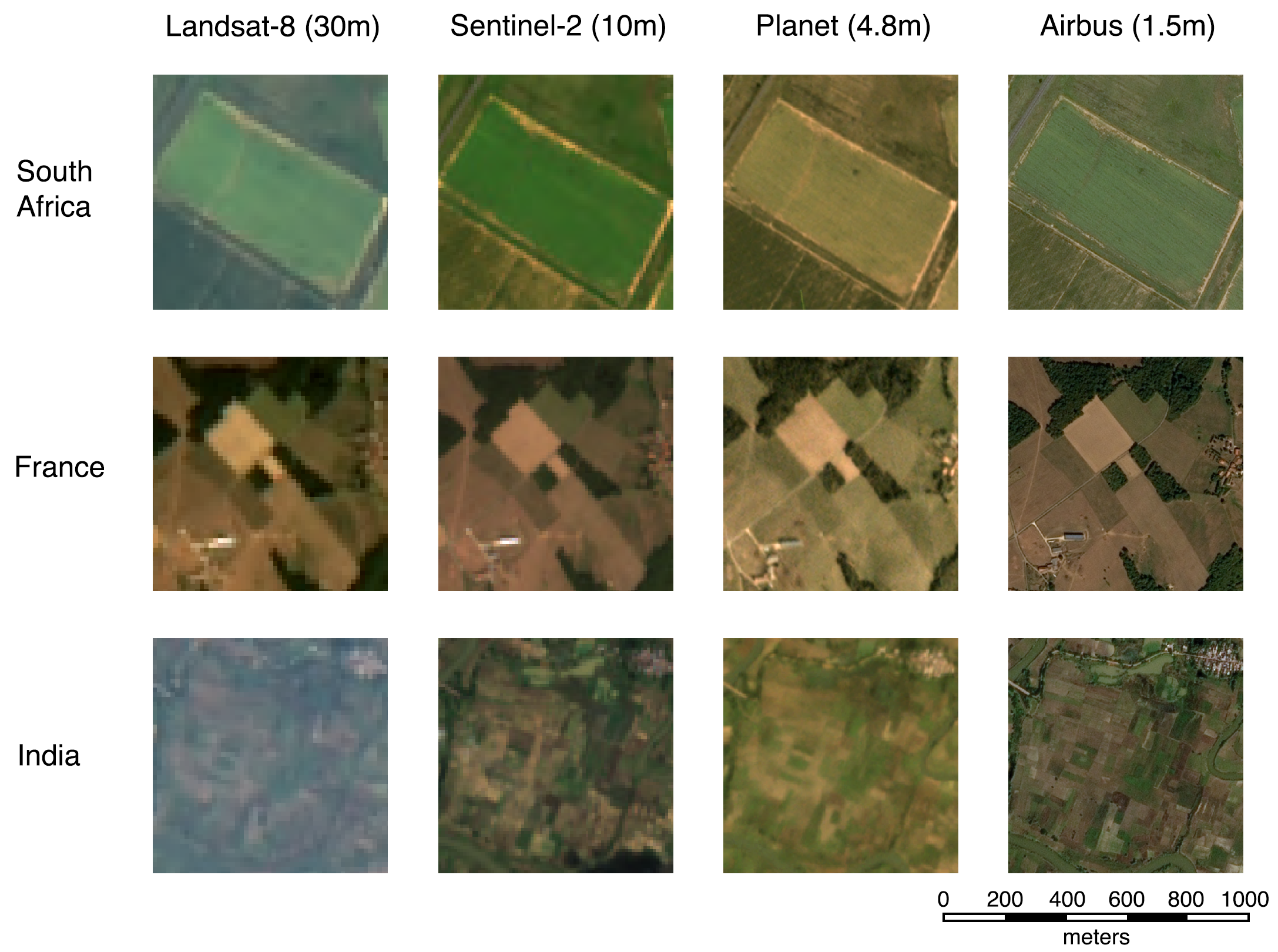}
	\caption{\textbf{Satellite images taken over agricultural areas in South Africa, France, and India.} From left to right, images are taken by Landsat-8 at 30m resolution, Sentinel-2 at 10m resolution, Planet's PlanetScope at 4.8m resolution, and Airbus SPOT at 1.5m resolution. All images in South Africa were taken in February 2020, France in September 2019, and India in October 2020. As fields get smaller, higher resolution satellites are needed to delineate field boundaries.}
	\label{fig:example_images}
\end{figure}

High-resolution satellite imagery and recent advances in computer vision offer opportunities for automated field boundary delineation at low cost.
Researchers have demonstrated that delineation can be automated in industrial agricultural systems like those in North America \cite{yan2016conterminous, yan2014automated, rahman2019crop}, Europe \cite{aung2020farm, masoud2020delineation, rydberg2001integrated}, and Australia \cite{waldner2021detect} using publicly-available Landsat or Sentinel-2 imagery.
In some cases, previous methods used large quantities of historical field boundaries and supervised machine learning to prove automated delineation is possible. For example, \citet{waldner2020deep} used a novel deep learning architecture to delineate fields accurately in South Africa using boundaries from a government agency \cite{daff2006field}.
In other cases, historical data was unavailable even in high-income countries, and researchers collected their own data for the study. \citet{waldner2021detect} obtained over 70,000 field labels in Australia and used them to train a deep learning model that then delineated 1.7 million fields throughout the Australian grains zone.
Still other methods circumvent the need for labeled boundaries by employing unsupervised methods. In the US, \citet{yan2016conterminous} used edge detection and active contour segmentation to delineate fields in the US.

These advances in industrial agricultural systems offer reasons to be optimistic about smallholder field delineation.
Indeed, a number of recent works have applied deep learning methods to smallholder field delineation with success. For example, \citet{persello2019delineation} used very high resolution satellite imagery and labels collected via ground surveys to train a convolutional neural network (CNN) to predict field contours in Nigeria and Mali. \citet{zhang2021automated} used multi-temporal Sentinel-2 images and a recurrent residual U-Net to delineate fields in Heilongjiang province, China.
However, gaps still exist between prior work and delivering large-scale, high-accuracy field boundaries. \citet{persello2019delineation} were constrained by highly localized field boundary labels to study areas of a few square kilometers, and the study produced broken field boundaries rather than field instances (i.e. individual fields, the ultimate goal). \citet{zhang2021automated} worked at a larger spatial scale but used Sentinel-2 imagery (10m resolution), which prevented small fields from being delineated.
In order for automated field delineation to extend to smallholder systems, a number of challenges must be overcome.

The first major challenge to smallholder field delineation is the availability of satellite images at high enough resolution to see field boundaries. Figure \ref{fig:example_images} compares the appearance of fields in South Africa, France, and India in imagery taken by Landsat-8, Sentinel-2, PlanetScope, and Airbus SPOT satellites. 
Field boundaries in South Africa and France can be seen clearly in Landsat-8 (30m) and Sentinel-2 (10m) imagery, but smallholder fields in India require PlanetScope (4.8m) or Airbus SPOT (1.5m) imagery to be resolved (Figure \ref{fig:example_images}).
Historically, very high resolution satellite imagery was expensive to access and available only for a fraction of the Earth's surface.
Only recently has the cost of access been reduced either via the launch of cubesats (PlanetScope) or user-friendly integration of very high resolution basemaps (Airbus SPOT) in cloud-based platforms like Descartes Labs. Geographic coverage has also expanded as more satellites have been launched and data storage has become cheaper.

The second major challenge is the lack of labeled data for training and validating models to delineate smallholder fields.
As previously mentioned, ground- or aerial survey-based field boundaries are scarce in low-income regions. This leaves manual annotation of satellite imagery as the most viable source of boundaries \cite{waldner2020deep, vlachopoulos2020delineation, north2019boundary}. 
Assuming the first challenge is overcome and one can access very high resolution imagery for generating labels, the small size of smallholder fields still poses a challenge: the smaller the average field size, the more fields there are to label in each satellite image.
For example, suppose we want to generate segmentation labels for 1000 images. 
In a region where the average field size is 1 hectare, at PlanetScope resolution (4.8m) a typical $256 \times 256$-pixel image would contain 150 fields. Labeling 1000 images would equal labeling 150,000 fields. The labor required to create fully-segmented labels that sample a representative fraction of a state, province, or country is therefore quite large, pointing to a need for more efficient ways of generating field boundary labels.

In this work, we explore how these two challenges can be overcome to accurately delineate field boundaries across India. We experiment with using both multi-temporal but lower-resolution PlanetScope imagery and very-high-resolution but single-date Airbus SPOT imagery as inputs to deep learning models. 
To overcome the lack of field boundary labels, we generate a new dataset of 10,000 manually-annotated field boundaries across India to train and validate models. Unlike the fully-segmented labels usually used to train  delineation models, the labels we created are what we call \textbf{partial labels}: only a fraction of the fields in each image are labeled, allowing fields from more locations across India to be sampled for the same labeling budget. Using partial labels to train a neural network is considered a ``weak supervision'' strategy \cite{wang2020weakly}.
Lastly, we test whether a model trained on field boundaries in France can reduce the number of labels required in India to achieve accurate delineation. In transfer learning terms, we use knowledge gained from delineating fields in a source domain---France---to facilitate delineation in a target domain---India. 

The main findings of the work are as follows.
\begin{enumerate}
\item We use an attention-based CNN \cite{diakogiannis2021looking} followed by watershed segmentation to delineate field instances accurately ($\text{F1-score}=0.93, \text{MCC}=0.65, \text{median IoU} = 0.86$) across India. Airbus imagery yields higher performance than PlanetScope imagery.
\item We show that fully-segmented labels are not necessary for training CNNs to delineate fields. By masking out unlabeled parts of an image, we successfully supervise field delineation with partial labels. This relieves the labeling burden when delineating fields in a new region.
\item In situations where the total number of fields that can be labeled is constant, we find that it is better to label a few fields in many images than many fields in a few images.
\item We show that downsampling satellite imagery in France to match France field sizes to India field sizes results in better model transfer from France to India. Even without seeing any fields in India, a model trained on downsampled France imagery can delineate fields in India moderately well ($\text{F1-score}=0.88, \text{MCC}=0.50, \text{median IoU} = 0.68$).
\item Pre-training a model to delineate fields in France improves performance in India substantially when the number of field labels in India is small. Put differently, up to $20\times$ fewer labels are needed in India to achieve the same performance if knowledge is transferred from France.
\end{enumerate}
These results show that, by efficiently collecting partial labels and combining high resolution satellite imagery with transfer learning, smallholder fields can be delineated at large geographic scales. In the Discussion, we summarize an approach to delineating fields in regions with no existing field boundary datasets and highlight settings where field delineation remains challenging. Our final contribution is the release of our dataset of 10,000 fields in India, along with the weights of the field delineation models.

\section{Datasets}
\label{sec:data}

In this section, we describe the satellite imagery and the assembly of country-wide field boundaries in India and France. We provide details on how we sampled imagery for model training and validation, which Planet and Airbus satellite imagery we used, how we generated 10,000 new field labels in India, and where field boundary data come from in France.

\subsection{Sampling locations for datasets}

\begin{figure}
	\centering
	\includegraphics[width=0.9\linewidth]{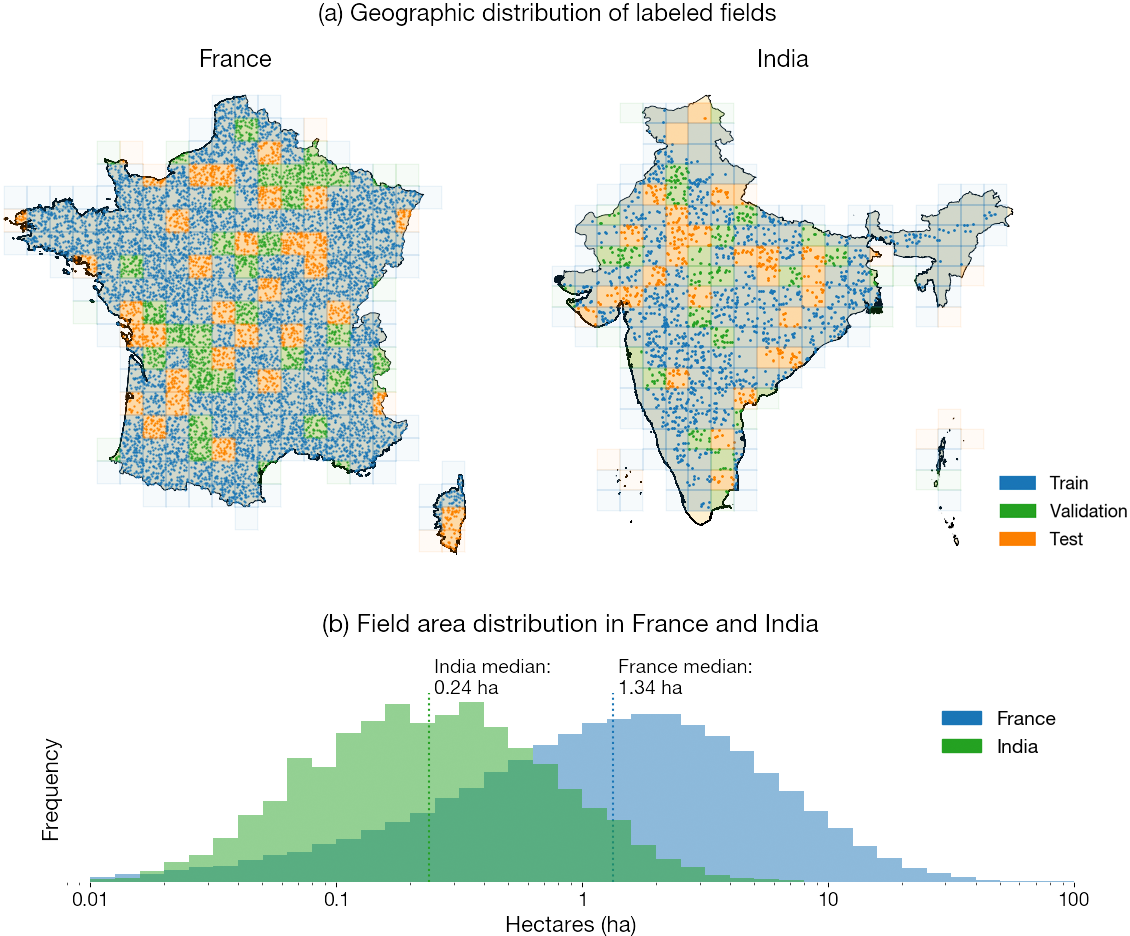}
	\caption{\textbf{Characteristics of France and India datasets.} (a) The locations of labeled fields are plotted as points on maps of France and India. We also visualize how images are split into training, validation, and test sets in geographic blocks. (b) Histograms of field areas on a log $x$-axis show that France fields are on average $5.6$ times larger than India fields.}
	\label{fig:datasets}
\end{figure}
\begin{table}[t]
\footnotesize
\centering
\begin{tabular}{x{0.15\linewidth} x{0.09\linewidth} x{0.09\linewidth} x{0.09\linewidth} x{0.11\linewidth} x{0.09\linewidth} x{0.09\linewidth} }
\toprule
\multirow{2}{*}{\textbf{Country}} & 
    \multicolumn{3}{x{0.33\linewidth}}{\textbf{Number of images}} &
    \multicolumn{3}{x{0.33\linewidth}}{\textbf{Number of fields}} \\
\cmidrule(lr){2-4} \cmidrule(lr){5-7}
 & 
    Train &
    Val &
    Test &
    Train &
    Val &
    Test \\
\midrule
France &
    6,759 &
    1,546 &
    1,568 &
    1,973,553 &
    459,512 &
    430,462 \\
\midrule
India &
    1,281 &
    300 &
    399 &
    6,421 &
    1,500 &
    1,996 \\
\bottomrule
\end{tabular}
\vspace{10pt}
\caption{\textbf{Number of images and fields in each split of the France and India datasets.} Training, validation, and test sets were split in geographic blocks, with the split being around 64\%-16\%-20\% train-val-test.}
\label{table:datasets}
\end{table}

\subsubsection{India}
\label{sec:india_sample}

To determine where in India to download satellite imagery and label fields, we used the Geo-Wiki dataset as a guide. Geo-Wiki contains a random sample of land surface locations globally that have been verified by crowdsourced volunteers to be cropland \cite{bayas2017global}. 
We manually inspected Airbus images centered at Geo-Wiki locations until we obtained a dataset of 2000 clear images for labeling. About a fifth of inspected images were removed due to the location not actually being cropland or the image being too low in contrast for fields to be delineated (Figure \ref{appendix:fig:geowiki_filtering}).
While some parts of India saw a greater chance for images to be rejected due to low contrast (e.g. along the coast, in very wet areas), the dataset still spans the entire country.

To split the images into training, validation, and test sets, we divided India into a $20 \times 20$ grid of cells (Figure \ref{fig:datasets}). We assigned 64\% of grid cells to the training set, 16\% to the validation set, and 20\% to the test set. All images that fell into a grid cell were assigned to the corresponding dataset split.
Splitting images along grid cells minimizes the chance that images across folds contain the same or very similar fields, thereby preventing classification metrics from being inflated due to leakage.

\subsubsection{France}
\label{sec:france_sample}

France is one of a handful of countries with publicly-available field boundary data. We use data in France to study the conditions needed for successful field delineation, and whether field delineation models trained in high-income regions can transfer to smallholder regions.

To construct a dataset of satellite imagery and field boundary labels in France, we first sampled 10,000 geographic coordinates at random from France's land surface. For each coordinate, we defined an image of $256\times 256$-pixels at 4.77m resolution (Planet imagery resolution) centered at the coordinate. The satellite image and field boundary ground truth were then obtained for each tile (see Sections \ref{sec:france_labels} and \ref{sec:data_planet}).

For model development and evaluation, we split the sampled locations into training, validation, and test sets. As in India, France was discretized into a $20\times 20$ grid of cells to minimize leakage among the splits. Each grid cell and all images that fell into it were placed in either the training, validation, or test set in a 64\%-16\%-20\% split (Figure \ref{fig:datasets}).

\subsection{Satellite imagery}

\subsubsection{Annual Airbus OneAtlas Basemap}
\label{sec:data_airbus}

The Airbus OneAtlas Basemap is a high-resolution map of Earth captured by the SPOT-6/7 satellites \cite{airbus_oneatlas}. 
The imagery is 1.5m resolution and captures four bands: red, green, blue, and near-infrared.
One basemap is created every year by stitching together hand-selected images with marginal cloud cover and seasonal consistency (meaning that adjacent imagery will be seasonally contiguous). The cloud cover target is less than 5\% in regular areas and less than 25\% in challenging areas; in practice, this means that most Airbus basemap images in India were taken during the dry season from October to March.
We used the Descartes Labs platform to access Airbus imagery.

Human annotators used Airbus imagery to label fields in India (see Section \ref{sec:data_india_labeling}). We also used Airbus imagery as a model input for field delineation. We tried both Airbus imagery at native resolution (1.5m) and down-sampled $3\times$ (4.5m) to compare against Planet imagery.

\subsubsection{Monthly PlanetScope Visual Basemaps}
\label{sec:data_planet}

We also explored using monthly PlanetScope Visual Basemaps, 
which have a ground sampling distance of 4.77 meters \cite{visual_basemaps}. A distinct Visual Basemap is generated globally for each month using Planet's proprietary ``best scene on top'' algorithm, which selects the highest quality imagery from Planet's catalog over the course of the month based on cloud cover and image sharpness.
A major advantage of Visual Basemaps is that they already chose the least cloudy image in an area, which is  helpful in the subtropics where clouds are frequent. However, a downside is that, unlike individual PlanetScope images, Visual Basemaps only offer RGB bands. Since differences in vegetation often appear in the NIR range, we expect future work that uses NIR bands to perform even better than the results shown in this paper.

In France, we downloaded April, July, and October 2019 Visual Basemaps for the 10,000 locations. These months were chosen to span the growing season; April and October in particular mark the beginning and end of the growing season, when the contrast between adjacent crop fields is likely to be largest due to variation in sowing and harvest dates. 

In India, we obtained Visual Basemaps for each location with labeled fields (see Section \ref{sec:data_india_labeling}). 
We sampled $512 \times 512$-pixel Planet tiles around each set of fields to achieve a larger effective dataset size; at training time, a random $256 \times 256$-pixel crop of the tile was taken, which allows the partial field labels to fall anywhere in the image (Figure \ref{appendix:fig:random_crops}). 
At each location, we downloaded Visual Basemaps for each month between August 2020 and July 2021. 
We later determined (Results Section \ref{sec:results:india}) which months' imagery was best suited for field segmentation.

\subsection{Field boundary labels}

\subsubsection{Creating field boundary labels in India}
\label{sec:data_india_labeling}

Since no large-scale, geographically representative dataset of crop fields exists in India to our knowledge, we employed human workers to annotate fields across the country. Existing field boundary datasets are assembled either through field surveys \cite{persello2019delineation}, farmer submissions \cite{france}, or manual inspection of aerial or high-resolution satellite imagery \cite{waldner2020deep}. Due to the expensive and time-consuming nature of field surveys, as well as a lack of digital infrastructure to query farmers about their field boundaries, we opted for the third strategy. 

In order to label fields in satellite imagery, we need imagery of high enough resolution for field boundaries to be clear to human annotators. Since neither Landsat, Sentinel-2, nor PlanetScope imagery are high enough resolution (Figure \ref{fig:example_images}), we chose to use Airbus OneAtlas Basemap imagery (1.5m resolution). 

At 2000 Geo-Wiki locations, we pulled Airbus images and asked human annotators to delineate the 5 fields that fall closest to the center of the image, which was marked with an asterisk. The number 5 was chosen to balance a wider geographic coverage (the fewer fields per image, the more images sampled, and the more of India's diverse geography is represented in the dataset) with the cost of labeling (the fewer fields per image, the more time annotators spend switching between images, and the higher cost per field). 
Our results showed that this was a good decision, as simulations in France yielded better models when models were trained on more images but fewer labels per image (Results Section \ref{sec:results:partial}). 
The final dataset consists of 10,000 fields in India, which were then split into training, validation, and test sets (Table \ref{table:datasets}).

For each sample in India, the label generated by human annotators was a binary raster at the same resolution as Airbus imagery (1.5 m). A pixel had a value of 1 if it was inside a field that the annotator decided to label, and 0 otherwise. We converted these labels into georeferenced polygons so that they can be paired with any remotely sensed data source.
Note that, despite being vectorized, the field polygons should be considered drawn to 1.5m accuracy.

Because fields in an Airbus image can have ambiguous boundaries---due to low image contrast, small field size, or image blurriness---we asked the annotators to label only field boundaries that are clear in the image. This may result in bias in the India field boundary dataset toward omitting fields that are too small or low contrast for humans to see in Airbus imagery.
The evaluation of field segmentation models in this study can therefore be understood as comparing how well the models perform relative to human interpretation of satellite imagery, rather than true fields.

\subsubsection{Registre Parcellaire Graphique}
\label{sec:france_labels}

Field boundary labels in France come from the Registre Parcellaire Graphique (RPG) \cite{france}.
The RPG is a georeferenced database of all agricultural fields in France that receive aid under the Common Agricultural Policy (CAP) of the European Union. 
An anonymized version of the dataset is released publicly each year, and we accessed this dataset at \url{https://www.data.gouv.fr/} \cite{france}. The entire 2019 database contains 9.6 million plots, each drawn to centimeter resolution \cite{asp2019rpg}. Although the RPG does not include farmland not receiving CAP aid, in reality 95+\% of French agricultural land is recorded in the RPG. 
Our 10,000 images sampled across France contain over 2.7 million fields from the RPG; Table \ref{table:datasets} describes how they are split into training, validation, and test sets.

\subsubsection{Rasterizing polygons to create labels}
\label{sec:raster_labels}

For each image, we rasterized the overlapping field polygons to create the following three labels to train our neural network. Using these three labels to supervise multi-task learning has previously been shown to outperform using only one label \cite[][Figure \ref{sec:results:partial}]{diakogiannis2020resunet}. 
\begin{itemize}
    \item The \textbf{extent label} describes whether each pixel in the image is inside a crop field. Pixels inside a crop field have value 1, while pixels outside have value 0.
    \item The \textbf{boundary label} describes whether each pixel is on the boundary of a field. Pixels on the boundary (two pixels thick) have value 1; other pixels have value 0.
    \item The \textbf{distance label} describes the distance of pixels inside fields to the nearest field boundary. Values are normalized by dividing each field's distances by the maximum distance within that field to the boundary. All values therefore fall between 0 and 1; pixels not inside fields take the value 0.
\end{itemize}

\section{Methods}
\label{sec:methods}

This section first describes the methods shared by all our experiments: field boundary detection using a neural network followed by field extraction via the watershed segmentation algorithm. Details are provided on neural network architecture and training hyperparameters, model training on partial labels, post-processing to derive instances from segmentation predictions, and evaluation metrics. Finally, we explain the different experiments conducted to optimize field delineation in India and transfer knowledge from France to India. 

\subsection{Neural network implementation}

The first step of our field delineation pipeline is to use an attention-based deep neural network architecture to detect field edges. The architecture, named FracTAL-ResUNet, was first proposed in \citet{waldner2021detect} and used to create production-grade field boundaries in Australia. We experimented with both the FracTAL-ResUNet, ResUNet-a \cite{diakogiannis2020resunet}, and regular U-Net \cite{ronneberger2015unet} architectures and found that the FracTAL-ResUNet had the best performance in France and India.

We briefly describe the FracTAL-ResUNet architecture and loss function here and refer the reader to \citet{diakogiannis2021looking} and \citet{waldner2021detect} for more details. A FracTAL-ResUNet has three main features, reflected in its name: (1) a self-attention layer called a FracTAL unit that is inserted into standard residual blocks, (2) skip-connections that combine the inputs and outputs of residual blocks (similar to in the canonical ResNet), and (3) an encoder-decoder architecture (similar to a U-Net). We used a FracTAL-ResUNet model with a depth of 6 and 32 filters in the first layer, following the architecture in \citet{waldner2021detect}.

The FracTAL-ResUNet is trained on a Tanimoto with complement loss. For ground truth labels $\mathbf{y}$, model predictions $\mathbf{\hat{y}}$, and a hyperparameter $d$, the Tanimoto similarity coefficient is defined as
\begin{equation}
    \mathcal{FT}^{d}\left( \mathbf{y}, \mathbf{\hat{y}} \right) = \frac{1}{2} \left( \mathcal{T}^{d} \left( \mathbf{y}, \mathbf{\hat{y}} \right) + \mathcal{T}^{d} \left( 1-\mathbf{y}, 1-\mathbf{\hat{y}} \right) \right)
\end{equation}
where
\begin{equation}
    \mathcal{T}^{d} \left( \mathbf{y}, \mathbf{\hat{y}} \right) = \frac{\mathbf{y} \cdot \mathbf{\hat{y}}}{2^{d} \left( \mathbf{y}^{2} + \mathbf{\hat{y}}^{2} \right) - \left( 2^{d+1} - 1\right) \mathbf{y} \cdot \mathbf{\hat{y}} }
\end{equation}
The Tanimoto similarity coefficient takes the value 1 when $\mathbf{\hat{y}}=\mathbf{y}$, i.e. the predictions are perfect. To maximize the similarity coefficient during training, we train to minimize the loss
\begin{equation}
    \mathcal{L}^{d} \left( \mathbf{y}, \mathbf{\hat{y}} \right) = 1 - \mathcal{FT}^{d}\left( \mathbf{y}, \mathbf{\hat{y}} \right)
\end{equation}
The Tanimoto loss was shown in previous work \cite{diakogiannis2020resunet} to result in more accurate boundary predictions than cross entropy loss.

We trained all models until convergence (usually at least 100 epochs) with a learning rate of 0.001. Depending on GPU memory constraints, the batch size for our experiments ranged from 4 to 8 (when experimental results are directly compared, batch size was constant across experiments). During training, data were augmented with the standard practice of random rotations and horizontal and vertical flips.

\subsection{Training on partial labels}

\begin{figure}
	\centering
	\includegraphics[width=0.8\linewidth]{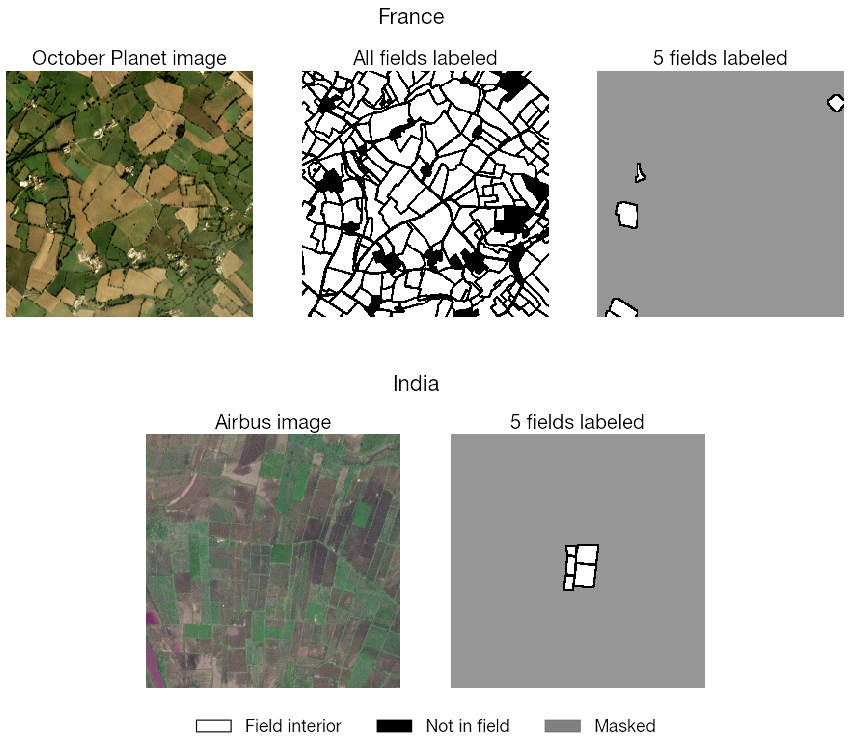}
	\caption{\textbf{Examples of partial labels.} Usually field delineation is trained on full labels, such as those available in France through the Registre Parcellaire Graphique. In this paper, we instead train on partial labels by masking out unlabeled parts of the image. Here we show examples of 5 fields labeled in France and India images, and denote masked pixels in gray.}
	\label{fig:sparse_label}
\end{figure}

Usually, labels used to train field delineation models \cite{waldner2020deep, waldner2021detect} are densely segmented, which means that every pixel in an image is annotated. 
Densely segmented labels are costly to create, so we instead generated partial labels, where only a fraction of the image is annotated. This type of training strategy---using imperfect or partial labels---falls under the machine learning sub-field of weakly supervised learning \cite{wang2020weakly}. We hypothesized that it is better to label fewer fields in more images than more fields in fewer images, with the rationale that sampling fields from more locations in a region leads to better generalization. Below we describe how we train models on partial labels and the experiment we conducted to test this hypothesis.

\paragraph{Masking out unlabeled areas}

Training a neural network on partial labels instead of fully segmented labels requires changes to the training procedure.
Instead of computing the loss at every pixel of the image, the loss is only computed at labeled fields and their boundaries. Unlabeled pixels are masked out (Figure \ref{fig:sparse_label}); the model's predictions are not evaluated there.
To implement masking, we created a binary mask for each partial label, where 1s correspond to labeled pixels and 0s correspond to unlabeled pixels. Before computing the loss, we multiply the predictions by the mask to set every pixel that falls outside of the mask to 0.

Similarly, at evaluation time, model performance is evaluated on the labeled fields of the validation and test set images but not on unlabeled pixels. 
The performance at labeled fields should generalize to unlabeled fields as long as fields were labeled at random.

Note that, since annotators were asked to annotate 5 fields per image, the models are never trained on large swaths of non-crop labels, only on fields and the boundaries between fields. This means that, unlike models trained on densely segmented labels, models trained on these partial labels need a cropland map in post-processing to mask out non-crop areas. Given the availability and increasing accuracy of global cropland maps, this modular approach to field segmentation should be feasible.

\paragraph{Varying fields labeled per image}

To optimize partial label collection, we simulated different annotation strategies using the fully-segmented France dataset.
We assume that the total labeling budget allows for 10,000 fields to be labeled, but the number of fields per image and number of images labeled can vary. On one extreme, 125 training images were sampled across France and then 80 field labels were sampled per image. On the other extreme, 5000 training images were sampled and only 2 field labels were sampled per image. We also compared 200 images with 50 fields per image, 500 images with 20 fields per image, 1000 images with 10 fields per image, and 2000 images with 5 fields per image (Figure \ref{fig:sparse_label}). All experiments were evaluated on the same validation and test set images, with partial labels of all fields in each image (i.e. models were evaluated on all fields in each image and their boundaries, but not on non-crop areas).

\subsection{Post-processing predictions to obtain field instances}

The deep learning model outputs field boundary detections but not separate crop fields. To obtain individual fields---also known as \textbf{instances}---we used a hierarchical watershed segmentation algorithm. Watershed segmentation is a region-based algorithm that operates on a grayscale image and treats it like a topographic map, with each pixel's brightness value representing its height. The algorithm separates objects within an image by seeding objects at local minima and then expanding them outward with increasing height (like water filling a drainage basin). Objects stop expanding when they reach the borders of other objects. Watershed segmentation has been shown to outperform other instance segmentation algorithms for field boundary delineation \cite{watkins2019comparison, waldner2020deep}. The specific implementation we used can be found in the Python package \texttt{higra} and is described in \citet{najman2013playing}. Both the field extent and field boundary predictions are required inputs to watershed segmentation. We tuned the algorithm's hyperparameters using validation set imagery.

\subsection{Evaluation metrics}

We evaluate two outputs from the field delineation pipeline: (1) pixel-level predictions output by the FracTAL-ResUNet and (2) field instances obtained after post-processing.

\paragraph{Semantic segmentation metrics}

Recall that the FracTAL-ResUNet outputs three predictions: field extent, field boundaries, and distance to field boundary. Each prediction is a raster of the same size and resolution as the satellite image input, with each pixel taking on values between 0 and 1. We evaluate only the extent prediction using overall accuracy, F1-score, and Matthews correlation coefficient.

The first metric is \textbf{overall accuracy (OA)}, which is the most commonly used classification metric. We convert the model's field extent prediction to a binary prediction by setting values $\ge 0.5$ to 1 and values $< 0.5$ to 0. OA is then defined as
\begin{equation}
    \textup{OA} = \frac{\textup{TP} + \textup{TN}}{\textup{TP} + \textup{FP} + \textup{FN} + \textup{TN}}
\end{equation}
where TP, TN, FP, and FN are the number of true positives, true negatives, false positives, and false negatives, respectively.
When all predictions are perfect, $\textup{OA} = 1$; when all predictions are incorrect, $\textup{OA} = 0$.

The second metric is \textbf{F1-score (F1)}, which is the harmonic mean of precision and recall. Mathematically, it is defined as
\begin{equation}
    \textup{F1} = 2 \times \frac{\textup{precision} \times \textup{recall}}{\textup{precision} + \textup{recall}} = \frac{\textup{TP}}{\textup{TP} + \frac{1}{2}\left(\textup{FP} + \textup{FN} \right)}
\end{equation}
The F1-score also ranges in value from 0 to 1, and requires both precision and recall to be high in order to be a high value.

The last and most discerning metric is the \textbf{Matthews correlation coefficient (MCC)}. The MCC for a set of binary classification predictions is defined as
\begin{equation}
    \textup{MCC} = \frac{\textup{TP} \times \textup{TN} - \textup{FP} \times \textup{FN}}{\sqrt{(\textup{TP} + \textup{FP})(\textup{TP} + \textup{FN})(\textup{TN} + \textup{FP})(\textup{TN} + \textup{FN})}}
\end{equation}
MCC measures the correlation between predictions and true labels, and is considered a more reliable metric than accuracy and F1 score  \cite{chicco2020advantages}. When the classifier is perfect, the value of MCC is 1, indicating perfect correlation. When the classifier is always wrong, the value of MCC is $-1$, indicating complete negative correlation. MCC is only high if the prediction obtains good results in all four confusion matrix quadrants (TP, TN, FP, FN), and it does not produce misleading high values on very imbalanced datasets (unlike accuracy and F1-score). MCC was used in previous field segmentation studies to assess model performance \cite{waldner2020deep, waldner2021detect}, and we will use it as our primary metric for assessing field boundary detection.

When training models for many epochs, we keep the weights at the epoch with the highest validation set MCC. In our results, we report accuracy, F1-score, and MCC for each experiment on the test set, which is held out and never seen during model training or hyperparameter tuning.

\paragraph{Instance segmentation metrics}

To evaluate the quality of predicted field instances after post-processing, we use the \textbf{Intersection over Union (IoU)} metric. IoU is a common metric for evaluating the accuracy of an object detector \cite{rezatofighi2019generalized}. Given a set of ground truth pixels and a set of predicted pixels for an object, IoU is defined as
    \begin{equation}
        \textup{IoU} = \frac{\textup{Area of overlap}}{\textup{Area of union}}
    \end{equation}
Perfect overlap yields an IoU of 1; no overlap, an IoU of 0.
For each ground truth field, we compute its IoU with the predicted field that has the largest overlap. When evaluating over all fields in a dataset, we compute the median IoU and the fraction of fields with IoU greater than $k\%$, which we denote as $\text{IoU}_{k}$. For example, $\text{IoU}_{50}$ is the fraction of fields where the overlap with a predicted field is at least half ($\ge 50\%$).

\subsection{Field delineation experiments}

\begin{figure}
	\centering
	\includegraphics[width=0.7\linewidth]{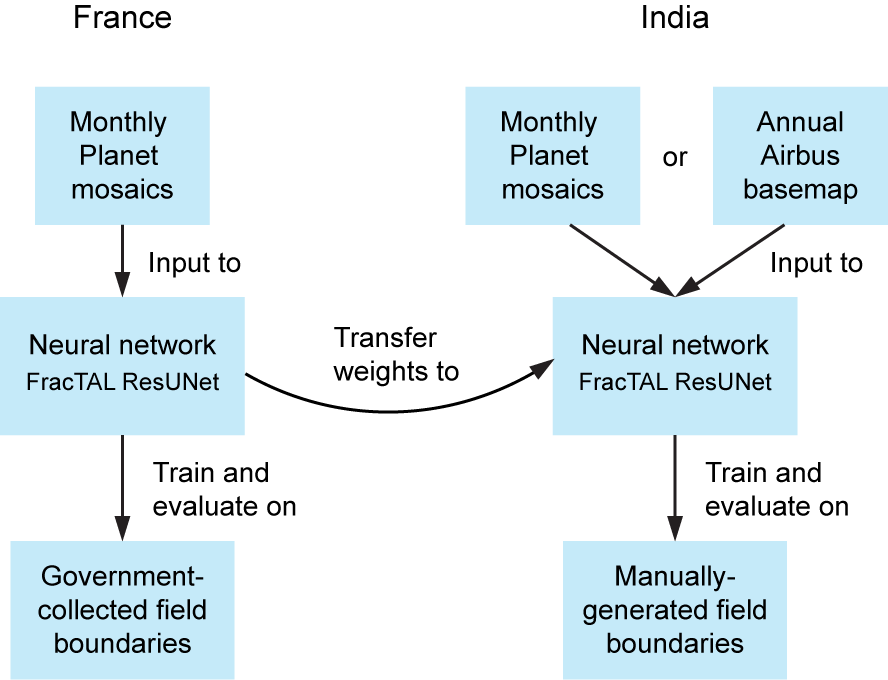}
	\caption{\textbf{Overview of field delineation experiments.} A model is first trained on imagery and field boundary labels in France before being transferred to delineate fields in India.}
	\label{fig:flowchart}
\end{figure}

We conduct experiments varying input imagery and degree of transfer learning to optimize field delineation in India. The experiments are summarized in Figure \ref{fig:flowchart} and described below.

\paragraph{PlanetScope vs. Airbus OneAtlas imagery}

In France, we delineated fields using Planet imagery since 4.8m is high enough resolution to clearly separate fields in France. In India, we conducted experiments with both Planet and Airbus imagery as input, since Indian fields can be extremely small. Airbus imagery was also downsampled by a factor of 3 to approximate Planet image resolution, in order to see whether differences between Airbus and Planet performance stemmed from Planet being lower resolution.

\paragraph{Combining multi-temporal imagery}

Since we obtained PlanetScope mosaics for multiple months, we experimented with two different ways of combining multi-temporal inputs. 
In the first method, the model is fed multiple months of imagery separately. For example, in France the number of samples in the dataset became 30,000 and each label was replicated in the dataset 3 times (for April, July, and October imagery). We then followed the approach in \citet{waldner2020deep} and averaged  model predictions across the months to obtain a \textbf{consensus} prediction.

The second approach was to stack multiple months of imagery and feed them as a single input to the neural network. In France, the input became a 9-band image where the first 3 bands were the April image, the second 3 bands were the July image, and the last 3 bands were the October image. We also experimented with shuffling the months in the stack at random for each sample. The rationale behind shuffling was that such a model would be more robust to India imagery looking different from France imagery, resulting in better performance when delineating fields in India.

\paragraph{Downsampling France imagery}

\begin{figure}
	\centering
	\includegraphics[width=1.0\linewidth]{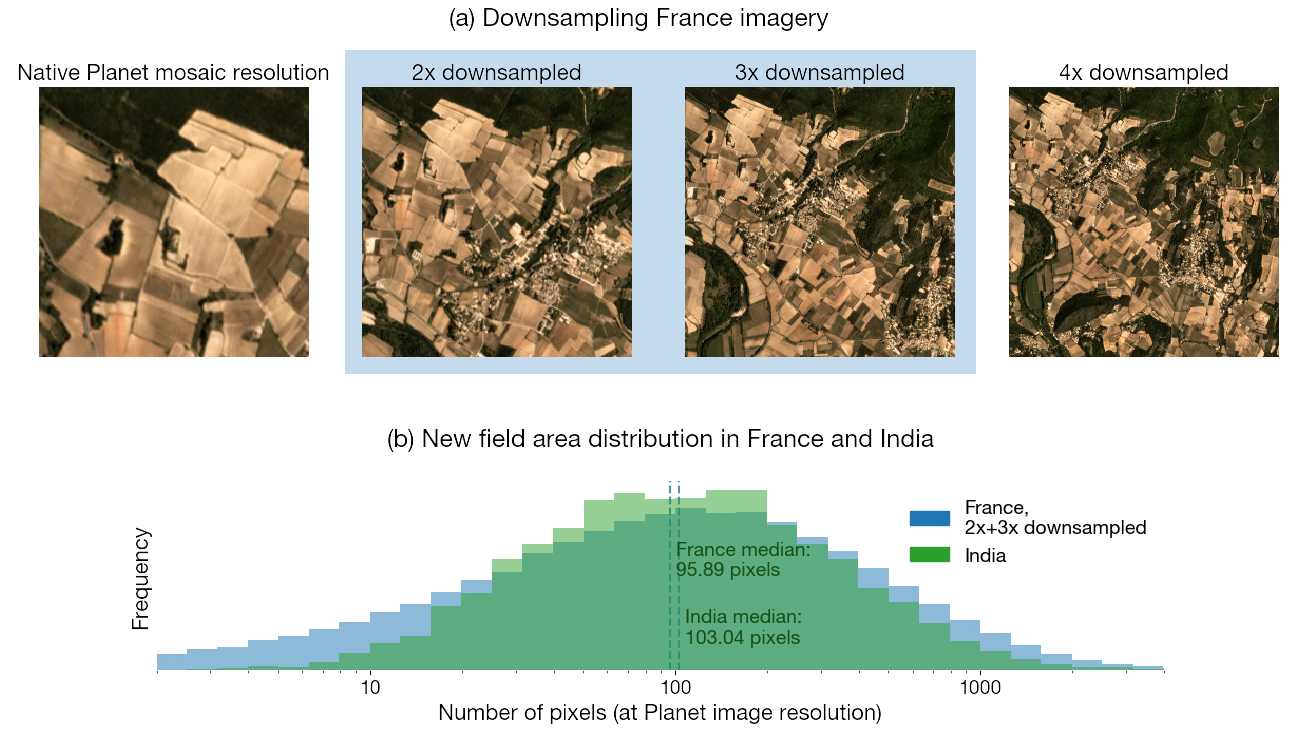}
	\caption{\textbf{Downsampling France imagery to match Indian field sizes.} (a) We transformed the France dataset by downsampling Planet imagery $2\times$ and $3\times$ and combining the two sets of images. (b) As a result, the transformed France field size distribution (as measured by Planet pixels per field) matches the India distribution more closely than the original (Figure \ref{fig:datasets}).} 
	\label{fig:downsampling}
\end{figure}

A major challenge of transferring field delineation models across geographies is that fields vary in size and shape in different parts of the world. Fields in France are on average 5.6$\times$ larger than fields in India (Figure~\ref{fig:datasets}). 

To improve model transfer from France to India, we downsampled each Planet image in France by $2\times$ and $3\times$, and trained a model on the union of these downsampled images. Note that a $2\times$ downsampling refers to downsampling $2\times$ along both the image height and width, which results in fields that are $4\times$ smaller in area. 
Downsampling causes France's field size distribution to overlap with India's field size distribution (Figure \ref{fig:downsampling}).

\paragraph{Training from scratch vs. transfer learning}

Knowledge transfer from high-income regions with ample field labels could facilitate field delineation in smallholder regions. In this work, France serves as the source domain and India as the target domain. To test knowledge transfer, the following models were used to delineate fields in India:
\begin{enumerate}
    \item A model trained on original-resolution Planet imagery and fully-segmented field labels in France
    \item A model trained on downsampled Planet imagery and fully-segmented field labels in France
    \item A model pre-trained on downsampled Planet imagery and fully-segmented field labels in France, then fine-tuned on Planet/Airbus imagery and partial field labels in India
    \item A model trained on Planet/Airbus imagery and partial field labels in India
\end{enumerate}
The first two models simulate what happens when we have zero labels in India, and comparing them reveals the benefit of downsampling France imagery to match India field sizes. Comparing model 2 to model 3 shows the benefit of fine-tuning on India labels, while comparing model 3 to model 4 shows the benefit of pre-training on France labels. 

We repeated the comparison of models 3 and 4 while varying the number of labels available in India for training/fine-tuning. Instead of using all 6400+ fields in the India training set, we restricted this number to 100, 200, 500, 1000, 2000, and 5000. For each training set size, a model was trained from scratch and compared to a model pre-trained in France and fine-tuned on the small training set.

\section{Results}
\label{sec:results}

\subsection{Field statistics in India}

We first describe the new field label dataset assembled as part of this study. The 10,000 fields in India ranged in size from 2 m$^{2}$ to 11 ha, with a median field size of 0.24 ha (Figure \ref{fig:datasets}). Although we use very high resolution satellite imagery, the smallest fields cannot be delineated even at Airbus SPOT resolution. Conversely, the largest fields can be delineated even with Sentinel-2 imagery. Most fields, however, fall in a range where they are suitable for delineation with PlanetScope or Airbus SPOT imagery.
For comparison, field sizes in France ranged from  3 m$^{2}$ to 2117 ha, with a median field size of 1.3 ha. France fields are therefore on average 5.6$\times$ larger than India fields. The smallest France fields are similar in size to the smallest India fields, while the biggest France fields are nearly 200$\times$ the size of the largest India fields.

Field sizes in India are not uniform across the country; fields are larger in northwestern and central India and smaller in eastern India (Figure \ref{appendix:fig:fieldsize}). This is consistent with findings from the Geo-Wiki database \cite{bayas2017global}, although our field sizes are more precise than the discretized bins in Geo-Wiki.

\subsection{Training on partial labels}
\label{sec:results:partial}

\begin{table}[t]
\footnotesize
\centering
\begin{tabular}{x{0.2\linewidth} x{0.25\linewidth} x{0.2\linewidth}}
\toprule
\textbf{Number of images} & 
    \textbf{Number of fields per image} & 
    \textbf{MCC} \\
\midrule
125 &
    80 &
    0.563 \\
200 &
    50 &
    0.585 \\
500 &
    20 &
    0.596 \\
1000 &
    10 &
    0.597 \\
2000 &
    5 &
    0.601 \\
5000 &
    2 &
    0.601 \\
\bottomrule
\end{tabular}
\vspace{10pt}
\caption{\textbf{Partial label experiment in France.} We simulate a situation in France where we are constrained to only collecting 10,000 field labels to answer whether it is better to collect full labels for a few images or partial labels for many images. We vary the number of images from 125 to 5,000 and the number of fields per image from 80 to 2, while keeping the total number of labeled fields constant at 10,000.}
\label{table:partial}
\end{table}

For the same labeling budget, labeling fewer fields per image across more images improved model performance. In our France simulation, training on 125 images with 80 fields per image yields an MCC of 0.563, while training on 5000 images with 2 fields per image yields an MCC of 0.601 (Table \ref{table:partial}). Between these two extremes, most of the advantage of having more images is realized by 500 images (MCC 0.596); adding more images beyond 500 only increases MCC slightly. These results suggest that not only are densely segmented labels not necessarily for supervising a deep learning model, but partial labels that allow more images to be labeled are actually preferable when labeling resources are constrained.

\subsection{PlanetScope vs. Airbus OneAtlas imagery}
\label{sec:results:india}

\begin{table}[t]
\scriptsize
\centering
\begin{tabular}{x{0.18\linewidth} x{0.05\linewidth} x{0.06\linewidth} x{0.06\linewidth} x{0.05\linewidth} x{0.06\linewidth} x{0.06\linewidth} x{0.05\linewidth} x{0.06\linewidth} x{0.06\linewidth} }
\toprule

\multirow{6}{*}{\textbf{Model}} &
    \multicolumn{3}{x{0.22\linewidth}}{\textbf{France}} &
    \multicolumn{6}{x{0.44\linewidth}}{\textbf{India}} \\ 
    \cmidrule(lr){2-4} \cmidrule(lr){5-10}
 &
    \multicolumn{3}{x{0.22\linewidth}}{\textbf{Planet imagery consensus\newline(Apr, Jul, Oct)}} & 
    \multicolumn{3}{x{0.22\linewidth}}{\textbf{Planet imagery consensus\newline(Oct, Dec, Feb)}} & 
    \multicolumn{3}{x{0.22\linewidth}}{\multirow{3}{*}{\textbf{Airbus imagery}}} \\
    \cmidrule(lr){2-4} \cmidrule(lr){5-7} \cmidrule(lr){8-10}
 & 
    \multicolumn{1}{x{0.05\linewidth}}{\textbf{OA}} &
    \multicolumn{1}{x{0.05\linewidth}}{\textbf{F1}} &
    \multicolumn{1}{x{0.061\linewidth}}{\textbf{MCC}} &
    \multicolumn{1}{x{0.05\linewidth}}{\textbf{OA}} &
    \multicolumn{1}{x{0.05\linewidth}}{\textbf{F1}} &
    \multicolumn{1}{x{0.061\linewidth}}{\textbf{MCC}} &
    \multicolumn{1}{x{0.05\linewidth}}{\textbf{OA}} &
    \multicolumn{1}{x{0.05\linewidth}}{\textbf{F1}} &
    \multicolumn{1}{x{0.061\linewidth}}{\textbf{MCC}} \\
\midrule

\scriptsize Trained in France (original Planet resolution) &
    0.91 &
    0.95 &
    0.48 &
    0.79 &
    0.87 &
    0.35 &
    0.77 &
    0.86 &
    0.29 \\
\midrule
\scriptsize Trained in France (downsampled Planet) &
    0.89 &
    0.93 &
    0.62 &
    0.76 &
    0.84 &
    0.39 &
    0.82 &
    0.88 &
    0.50 \\
\midrule
\scriptsize Pre-trained in France, fine-tuned in India &
    --- &
    --- &
    --- &
    0.82 &
    0.89 &
    0.52 &
    \textbf{0.89} &
    \textbf{0.93} &
    \textbf{0.65} \\
\midrule
\scriptsize Trained from scratch in India &
    --- &
    --- &
    --- &
    0.81 &
    0.88 &
    0.48 &
    0.88 &
    0.92 &
    0.64 \\
\bottomrule
\end{tabular}
\vspace{10pt}
\caption{\textbf{Pixel-level assessment of field delineation in France and India.} Table columns show results using a 3-month Planet image consensus in France and results using both the Planet consensus and Airbus imagery in India. Each row varies the amount of knowledge transfer from France and degree of additional training in India. The reported metrics are overall accuracy (OA), F1-score (F1) and Matthews correlation coefficient (MCC).}
\label{table:overview}
\end{table}
\begin{table}[t]
\scriptsize
\centering
\begin{tabular}{x{0.18\linewidth} x{0.09\linewidth} x{0.09\linewidth} x{0.09\linewidth} x{0.09\linewidth} x{0.09\linewidth} x{0.09\linewidth}}
\toprule

\multirow{6}{*}{\textbf{Model}} &
    \multicolumn{2}{x{0.22\linewidth}}{\textbf{France}} &
    \multicolumn{4}{x{0.44\linewidth}}{\textbf{India}} \\ 
    \cmidrule(lr){2-3} \cmidrule(lr){4-7}
 &
    \multicolumn{2}{x{0.22\linewidth}}{\textbf{Planet imagery consensus\newline(Apr, Jul, Oct)}} & 
    \multicolumn{2}{x{0.22\linewidth}}{\textbf{Planet imagery consensus\newline(Oct, Dec, Feb)}} & 
    \multicolumn{2}{x{0.22\linewidth}}{\multirow{3}{*}{\textbf{Airbus imagery}}} \\
    \cmidrule(lr){2-3} \cmidrule(lr){4-5} \cmidrule(lr){6-7}
 & 
    \multicolumn{1}{x{0.09\linewidth}}{\textbf{Median IoU}} &
    \multicolumn{1}{x{0.09\linewidth}}{\textbf{$\text{IoU}_{50}$}} &
    \multicolumn{1}{x{0.09\linewidth}}{\textbf{Median IoU}} &
    \multicolumn{1}{x{0.09\linewidth}}{\textbf{$\text{IoU}_{50}$}} &
    \multicolumn{1}{x{0.09\linewidth}}{\textbf{Median IoU}} &
    \multicolumn{1}{x{0.09\linewidth}}{\textbf{$\text{IoU}_{50}$}} \\
\midrule

\scriptsize Trained in France (original Planet resolution) &
    0.68 &
    0.64 &
    0.53 &
    0.53 &
    0.39 &
    0.41 \\
\midrule
\scriptsize Trained in France (downsampled Planet) &
    0.63 &
    0.61 &
    0.59 &
    0.59 &
    0.68 &
    0.67 \\
\midrule
\scriptsize Pre-trained in France, fine-tuned in India &
    --- &
    --- &
    0.70 &
    0.69 &
    \textbf{0.86} &
    \textbf{0.85} \\
\midrule
\scriptsize Trained from scratch in India &
    --- &
    --- &
    0.72 &
    0.69 &
    0.85 &
    0.82 \\
\bottomrule
\end{tabular}
\vspace{10pt}
\caption{\textbf{Instance-level assessment of field delineation in France and India.} Table columns show results using a 3-month Planet image consensus in France and results using both the Planet consensus and Airbus imagery in India. Each row varies the amount of knowledge transfer from France and degree of additional training in India. The reported metrics are median Intersection over Union (IoU) and the fraction of fields with IoU over 50\% ($\text{IoU}_{50}$).}
\label{table:iou}
\end{table}
\begin{figure}
	\centering
	\includegraphics[width=1.0\linewidth]{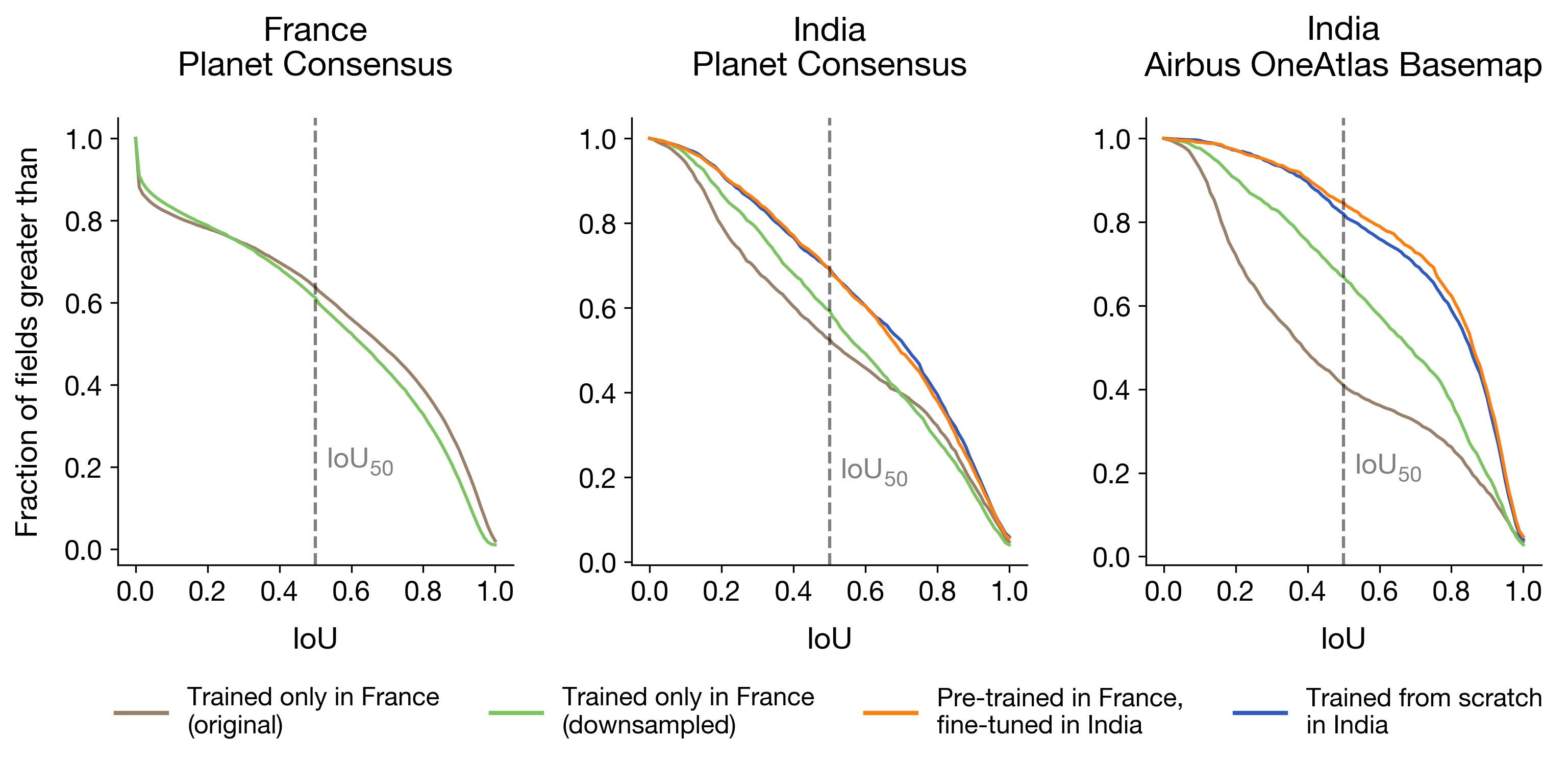}
	\caption{\textbf{IoU curves in France and India.} The $x$-axis shows the intersection-over-union (IoU) metric, and the  $y$-axis shows the fraction of fields in the test set with an IoU above a threshold. Results are shown for four models: trained only on France data (original resolution), trained only on France data (downsampled), pre-trained on France data (downsampled) and fine-tuned in India, and trained from scratch in India. Curves toward the upper right correspond to higher performance.}
	\label{fig:IoU_curves}
\end{figure}

In France, combining 3 months of Planet imagery by stacking them into one 9-banded image performed better than taking the consensus of 3 separate predictions (MCC 0.64 vs. 0.62, respectively). However, in India, the opposite was true; stacking resulted in slightly lower performance than taking the consensus (MCC 0.48 vs. 0.52, respectively). 
One possible explanation is that stacking requires more model parameters to be learned, which is worse for a smaller dataset like that in India but better for a larger dataset like that in France. Since our goal is to delineate fields in India, for the remainder of the results we combine Planet imagery by taking the consensus prediction across months.

When we trained on monthly Planet images separately, the three months with the highest average MCC were October, December, and February. We therefore averaged these three months to obtain a Planet consensus (Figure \ref{fig:example_predictions}). We also tried a 6-month consensus (adding August, April, and June) as well as a 12-month consensus (August 2020 to July 2021). However, adding more months decreased the consensus performance. The best Planet image-based result in India was an MCC of 0.52 and a median IoU of 0.70 (Table \ref{table:overview}, Table \ref{table:iou}). 

The results using Airbus outperformed results using Planet; the best Airbus image-based model in India achieved an MCC of 0.65 and a median IoU of 0.86. 
Surprisingly, using $3\times$ downsampled Airbus imagery resulted in better performance than using original-resolution Airbus imagery; the results shown in this section are therefore for $3\times$ downsampled Airbus. Even though the resolution of Planet and Airbus imagery are comparable (4.8m vs. 4.5m), Airbus performance is much higher (Tables \ref{table:overview} and \ref{table:iou}, Figure \ref{fig:IoU_curves}). This could be attributed to a combination of the sharper focus of Airbus imagery and bias in favor of Airbus due to it being the imagery source for label generation. Nonetheless, the Airbus results indicate that the FracTAL-ResUNet model recreated human field delineation to high accuracy.

Airbus predictions are more confident than Planet predictions (i.e. there are more values closer to 0 and 1 rather than near 0.5; Figure \ref{fig:example_predictions}); this could be explained by the Planet prediction being a consensus of 3 images, which brings predictions closer to 0.5 when they disagree across months.

\subsection{Transferring models trained in France to India}

\begin{figure}
	\centering
	\includegraphics[width=1.0\linewidth]{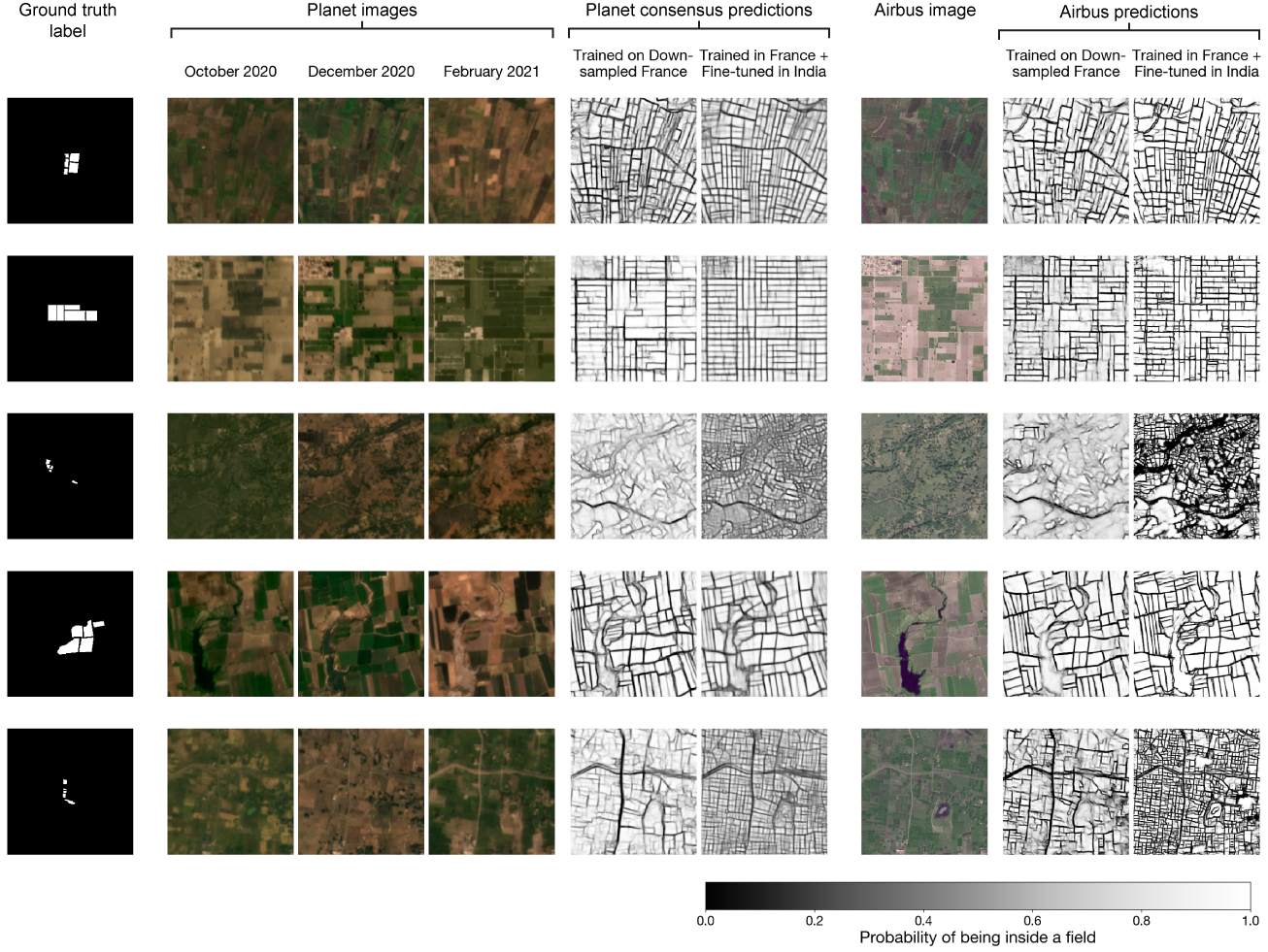}
	\caption{\textbf{Example model predictions in India.} We show the ground truth label, Planet Visual Basemap images for 3 months, Airbus OneAtlas Basemap images, and predictions using either Planet or Airbus images of models trained only in France and models trained in France then fine-tuned in India.}
	\label{fig:example_predictions}
\end{figure}

The rows of Table \ref{table:overview} and Table \ref{table:iou} show the performance of models trained only in France, only in India, and first in France and then in India. 
Since fields in France are on average 5.6$\times$ larger than fields in India, the model trained on original resolution Planet imagery is shown bigger fields during training than those it is asked to delineate in India. As a result, its pixel-level accuracy in India is quite low, with an MCC of 0.35 on Planet imagery and 0.29 on Airbus imagery (Table \ref{table:overview}). Field instance recovery is also poor, with a median IoU of 0.53 and 0.39, respectively (Table \ref{table:iou}).

In contrast, the model trained on downsampled France imagery performs better, with an MCC of 0.39 on Planet imagery and 0.50 on Airbus imagery, and corresponding median IoUs of 0.59 and 0.68. The performance on Airbus imagery is especially impressive, since the transfer is across sensors as well as geographies. Visualizing example predictions by models trained only on downsampled France data confirms that many field boundaries are correctly identified (Figure \ref{fig:example_predictions}).

Fine-tuning the models on India field labels improves performance substantially (second and third rows of Tables \ref{table:overview} and \ref{table:iou}). For the model using Planet imagery, MCC improves from 0.39 to 0.52 (median IoU from 0.59 to 0.70), while for the model using Airbus imagery MCC improves from 0.50 to 0.65 (median IoU from 0.68 to 0.86). Example predictions show that the improvement is especially dramatic for very small fields (Figure \ref{fig:example_predictions}). Comparing the third row to the fourth row, we find that training on India labels from scratch yields similar results to transfer learning from France. Training from scratch on Planet imagery yielded an MCC of 0.48 (median IoU 0.72), while training from scratch on Airbus imagery yielded an MCC of 0.64 (median IoU 0.85).

\begin{figure}
	\centering
	\includegraphics[width=1.0\linewidth]{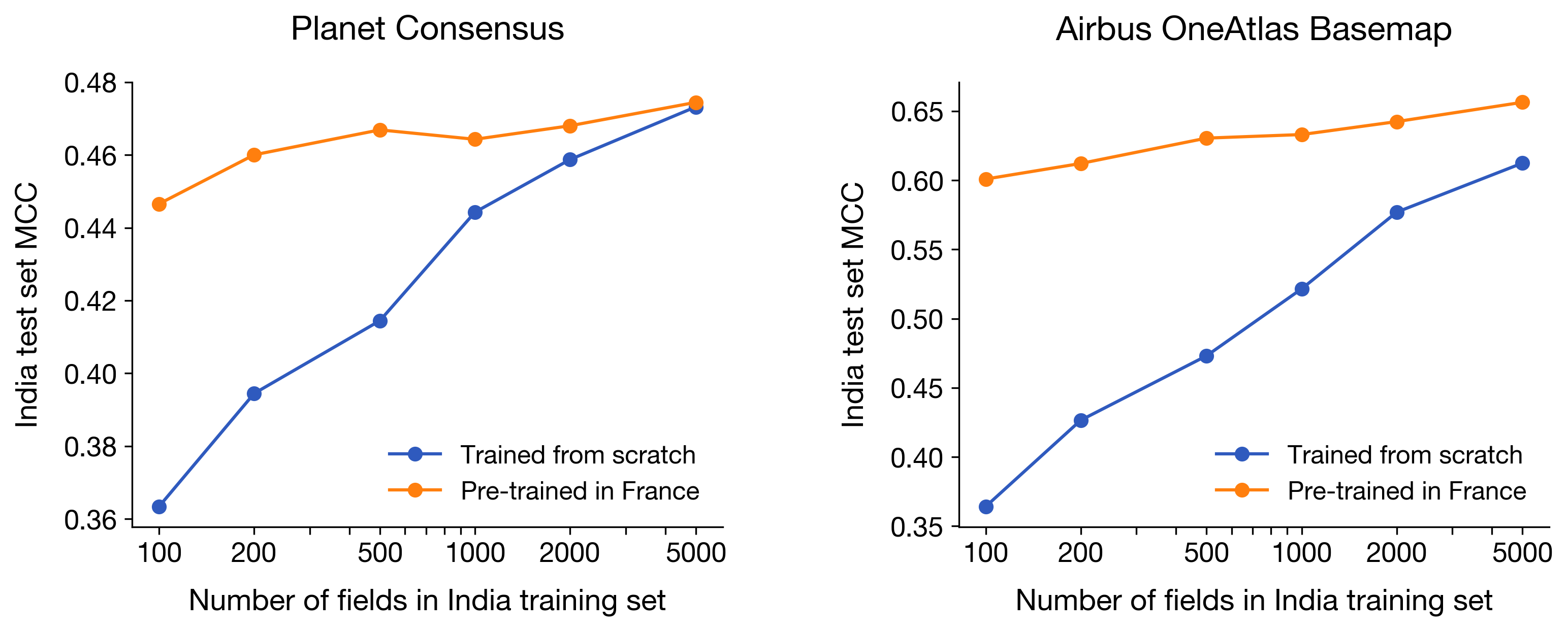}
	\caption{\textbf{The effect of pre-training a model in France.} We show the field delineation performance in India (measured by MCC) as we vary the number of labeled fields in the India training set for (left) a 3-month Planet consensus and (right) Airbus OneAtlas imagery. As the number of training fields decreases, pre-training in France before fine-tuning on India fields confers more of a benefit over training a model from scratch on the India fields.}
	\label{fig:dataset_size}
\end{figure}

At smaller training set sizes, however, the difference between transfer learning from France and training from scratch in India becomes substantial (Figure \ref{fig:dataset_size}). 
With 5000 training labels, the difference between pre-training in France and training from scratch is small (MCC of 0.474 vs. 0.473 for Planet, 0.66 vs. 0.61 for Airbus). However, at 100 labels, this difference becomes large, at 0.45 vs. 0.36 for Planet and 0.60 vs. 0.36 for Airbus. Indeed, the model pre-trained on France data and fine-tuned on 100 India labels performs better than the model trained on 1000 India labels from scratch when using Planet imagery and 2000 India labels from scratch when using Airbus imagery. With pre-training in France, the decrease in performance as dataset size diminishes is quite gradual. A mere 100 fields in India is enough to raise MCC of the Airbus model from 0.50 (directly applying a model trained in France) to 0.60 (fine-tuning on 100 fields in India). When datasets are small, pre-training in France reduces the number of India field labels needed to achieve a given performance level by as much as 20$\times$.

\begin{figure}
	\centering
	\includegraphics[width=1.0\linewidth]{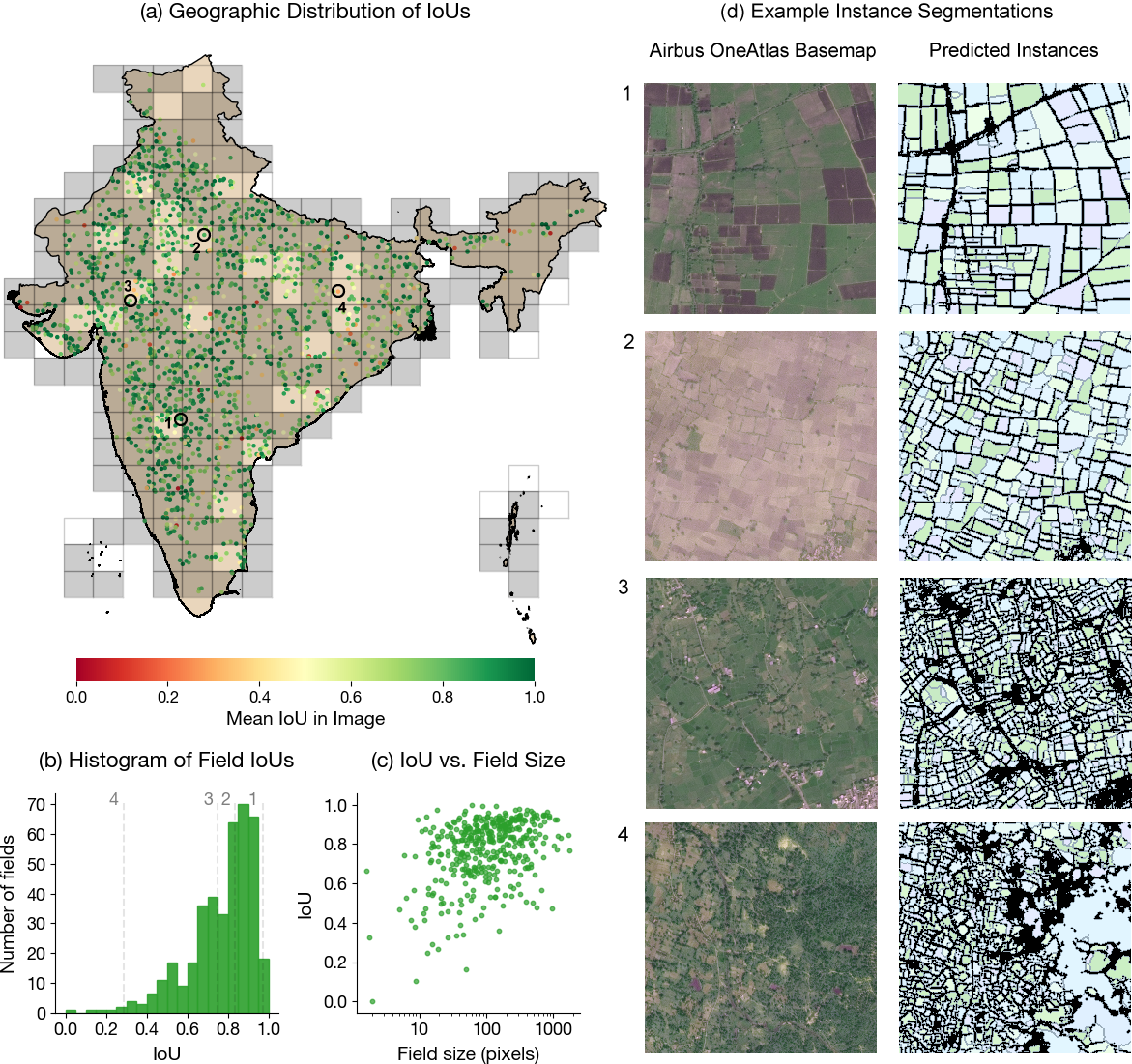}
	\caption{\textbf{Visualization of Airbus-based instance segmentation results.} (a) Mean IoU is plotted for each test set image over a map of India. Training and validation set grid cells are grayed out. (b) Histogram of field-level IoUs, with dashed lines showing the mean IoU of four example images in (d). (c) IoU and field size have a weak positive correlation. (d) Four example Airbus images and their corresponding instance segmentation results. The images are sorted from highest mean IoU (top) to lowest mean IoU (bottom).}
	\label{fig:overview}
\end{figure}

To better understand the factors that affect field delineation, we visualize the geographic distribution of IoU and the relationship between IoU and field size (Figure \ref{fig:overview}). 
We see that, while performance is high across India, IoUs are lower on average in east India than in west and central India. There is also a weak positive association between field size and IoU. Although the sample size is small, it appears that it may be difficult to achieve IoUs higher than $0.8$ when fields span fewer than $\sim 10$ pixels in an image. Figure \ref{fig:overview}d also shows example instance predictions for four images with average IoUs of 0.97, 0.84, 0.75, and 0.29. Images with medium-to-large field size and high contrast between fields result in accurate field delineation; meanwhile, small field size, low image contrast, and arid growing conditions appear to contribute to model under-segmentation (Figure \ref{appendix:fig:errors}).

\section{Discussion}
\label{sec:discussion}


Few datasets of field boundaries exist in smallholder agricultural systems. In this study, we assembled a dataset of 10,000 fields in India and used high resolution satellite imagery, transfer learning, and weak supervision with partial labels to automatically delineate smallholder fields. Building upon prior work \cite{waldner2020deep, waldner2021detect}, our methods achieve high performance (best model $\text{MCC} = 0.65, \text{median IoU} = 0.86$) through (1) access to very high resolution satellite imagery and (2) the use of partial field labels that relieve the labeling burden while enabling a large number of locations across the country to be sampled. In fact, taking a model naively trained in France and applied as-is in India as a benchmark, our method improves field-level identification by over two-fold. Alone, data augmentation techniques to mimic field sizes in India with France data improve IoU by up to 74\%. Our method also yields better results than a model trained from scratch only with India data for small sample sizes; pre-training reduce the number of India labels by as much as $20\times$.

Our findings support the following approach for delineating fields in a target region with no existing field boundary dataset: 
\begin{enumerate}
	\item Pre-train a FracTAL-ResUNet neural network on source-region imagery and field boundaries.
	\item Obtain remote sensing imagery of the appropriate resolution to resolve fields accurately in the target region.
	\item Create partial labels for a representative sample of fields across the target region.
	\item Fine-tune the neural network on a training set of labels in the target region.
	\item Evaluate the neural network on a test set of labels in the target region. Repeat Steps 3 and 4 until model performance is satisfactory.
\end{enumerate}
On a first pass, the user can skip Step 4 and generate just enough labels in Step 3 to evaluate a model trained only in the source region, since sometimes such a model may already yield good results. Should results not be satisfactory, one can iterate on Steps 3 and 4 until performance improves sufficiently. 

The experiments with transfer learning suggest that Step 1 (pre-training on source-region data) will greatly reduce the number of target-region labels required from Step 3 to achieve the user's desired performance level. This implies that, for the same labeling budget, fields across a much larger geographic area can be delineated with transfer learning than without. The methods described in this paper can therefore greatly speed up the development of field boundary datasets at large scales. There is also a ``network effect'' for field boundaries---once a region similar to the target region has labels, the target region needs many fewer labels for successful field delineation. The 10,000 India fields labeled for this work can also be used for pre-training and can therefore facilitate field delineation in other smallholder regions. Once all agro-ecological zones and field shapes have sizable field boundary datasets, future field delineation should progress quickly due to transfer learning.

Some of the above steps could vary depending on characteristics of the target region. For example, pre-training could be done with France data, another source region's data, or with labels pooled from multiple regions. Greater similarity between the source region and the target region should yield a larger performance improvement from pre-training; therefore, the source region should be chosen to match the agro-ecological zone and field shapes of the target region as much as possible. Similarly, the downsampling factor for pre-training imagery should be chosen to match the effective field size in the source region to the field size in the target region.

While our method expands the geographies around the world where field boundaries can be created, a few barriers may still prevent successful field delineation in some regions or by some users. First, high-resolution satellite imagery may not be available due to either lack of user access or cloudiness in the region of interest. Clouds are especially likely to pose a challenge in the tropics. In such settings, aerial imagery could be an alternative source of inputs, since airplanes and drones fly beneath clouds and aerial imagery has been used to delineate fields in the past \cite{vlachopoulos2020delineation}. We leave an investigation of how well field delineation models can transfer between satellite and aerial imagery to future work.

Second, in some regions it may be difficult even for humans to annotate fields. We encountered examples of this in India due to low contrast between fields in imagery (Figure \ref{appendix:fig:geowiki_filtering}).
One solution to low contrast could be to reference multiple satellite or aerial images during labeling to increase the chance that fields will appear distinct from their neighbors in at least one image.
Alternatively, satellite or aerial imagery-based annotations could be replaced by field survey-based field boundaries. Both alternate approaches---especially the latter---would increase labeling time and costs. Furthermore, we would expect a model using remote sensing imagery to perform less well in settings that require surveys to clarify field boundaries.
However, where fields can be distinguished in satellite imagery by humans, we expect the above pipeline to work well.

\section{Conclusion}
\label{sec:conclusion}

To date, crop field delineation in smallholder systems has been challenged by the low accessibility of high resolution satellite imagery and a lack of ground truth labels for model training and validation. This work combines high resolution Planet and Airbus imagery, a dataset of 10,000 new field instances, state-of-the-art deep learning, weak supervision with partial labels, augmentation of France imagery, and transfer learning to automatically delineate crop fields across India. 

First, human workers labeled 5 fields per image across 2,000 images from throughout India. 
By training models on partial labels instead of densely-segmented labels, we reduced the labeling burden and sampled a broader diversity of landscapes across India. 
Second, a neural network was pre-trained on field boundary data in France, where satellite imagery is also augmented to appear more like imagery in India. 
Lastly, the model was fine-tuned to predict partial field boundaries in India using either Planet or Airbus imagery.
Results show that accurate field delineation can be achieved with Airbus imagery, with our best model obtaining a pixel-level MCC of 0.65 and instance-level $\text{median IoU}$ of 0.86---more than double what a model trained naively in France and applied in India as-is achieved. Further experiments showed that pre-training in France reduces up to 20$\times$ the quantity of field labels needed in India to achieve a particular performance level.

Our method offers a scalable approach to delineating fields in regions lacking field boundary datasets. We release the dataset of 10,000 India field boundaries and trained model weights to the community\todo{URL will be added at publication}, with the goal of facilitating further method development and applications like crop type mapping, yield mapping, and digital agriculture in under-resourced regions of the world. 
\newpage
\section*{Acknowledgements}

This work was supported by the NASA Harvest Consortium
(NASA Applied Sciences Grant No. 80NSSC17K0652, sub-award 54308-Z6059203 to DBL) and a Google Cloud Credit Award from Stanford's Institute for Human-Centered Artificial Intelligence. Work by SW was partially supported by the Ciriacy-Wantrup Postdoctoral Fellowship at the University of California, Berkeley. We thank Descartes Labs for improving researchers' access to high resolution satellite imagery, and Rose Rustowicz in particular for helping us learn how to use the platform.
\appendix

\section{Appendix}
\label{sec:appendix}
\setcounter{figure}{0}

\begin{figure}[ht]
	\centering
	\includegraphics[width=0.7\linewidth]{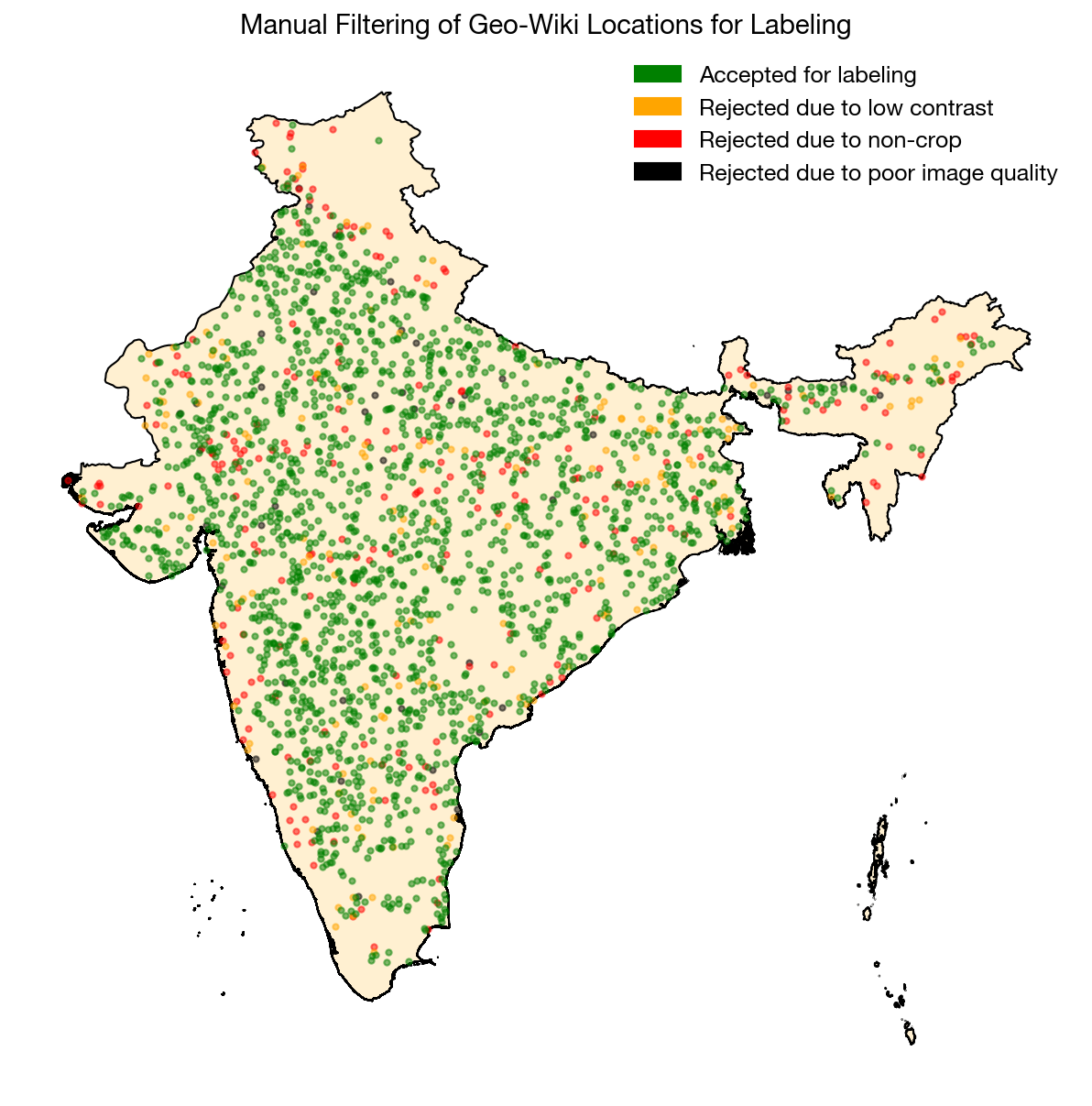}
	\caption{\textbf{Geographic distribution of locations accepted and rejected for field boundary labeling.} The authors inspected 2446 Airbus images over Geo-Wiki locations and approved 2000 of them for workers to annotate. The reasons for rejecting images were: image was too low contrast for labeling, image contained no crop or very little crop area, and image was low in quality.}
	\label{appendix:fig:geowiki_filtering}
\end{figure}
\begin{figure}[h]
	\centering
	\includegraphics[width=0.85\linewidth]{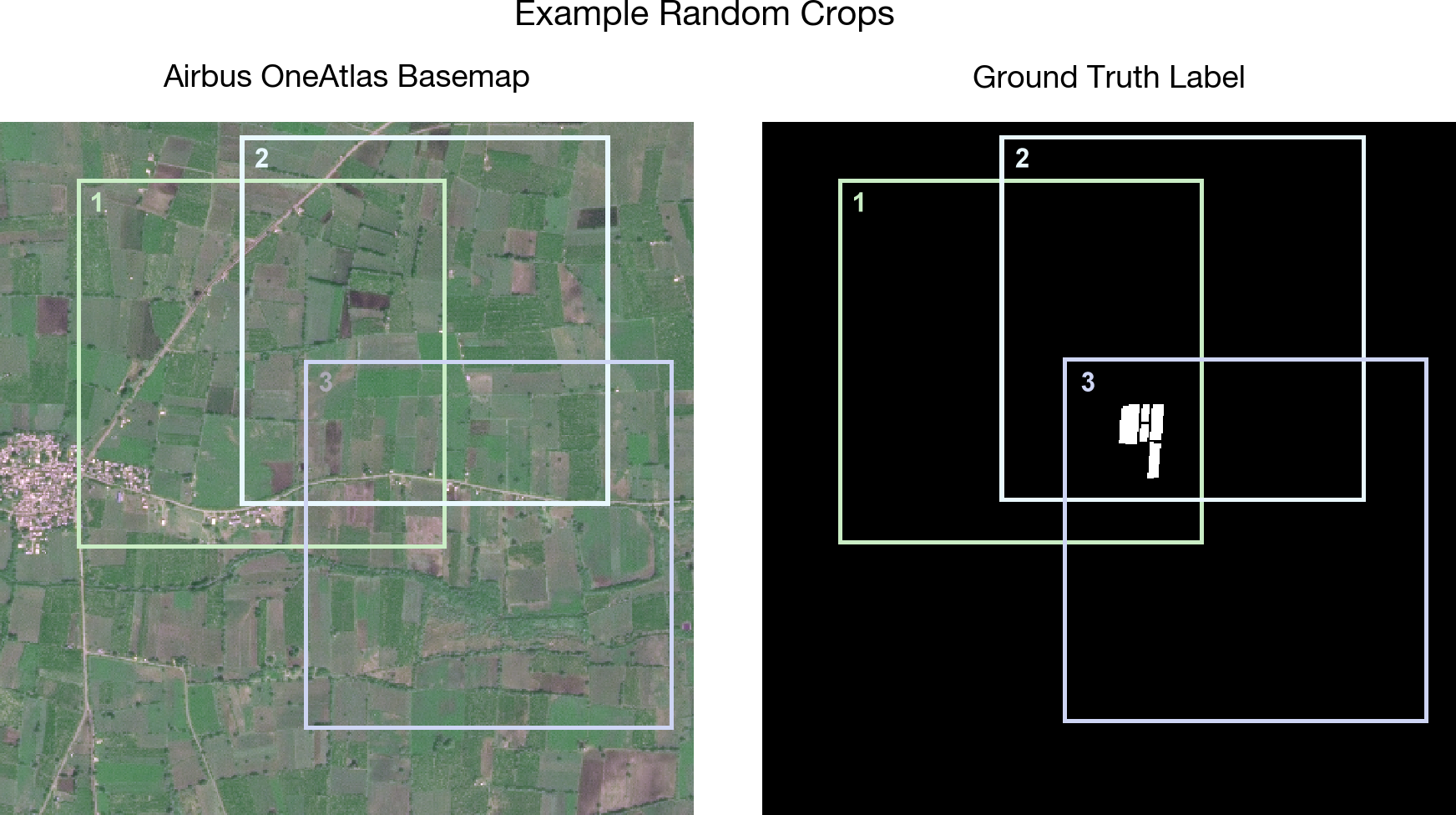}
	\caption{\textbf{Random image crops to increase effective dataset size.} We downloaded $512 \times 512$ pixel images around the labeled fields in India; at training time, a random crop of the image was taken. Compared to downloading $256 \times 256$ pixel images, this allows the model to see more diverse images, expanding the effective dataset size and delaying overfitting.}
	\label{appendix:fig:random_crops}
\end{figure}
\begin{figure}[h]
	\centering
	\includegraphics[width=0.85\linewidth]{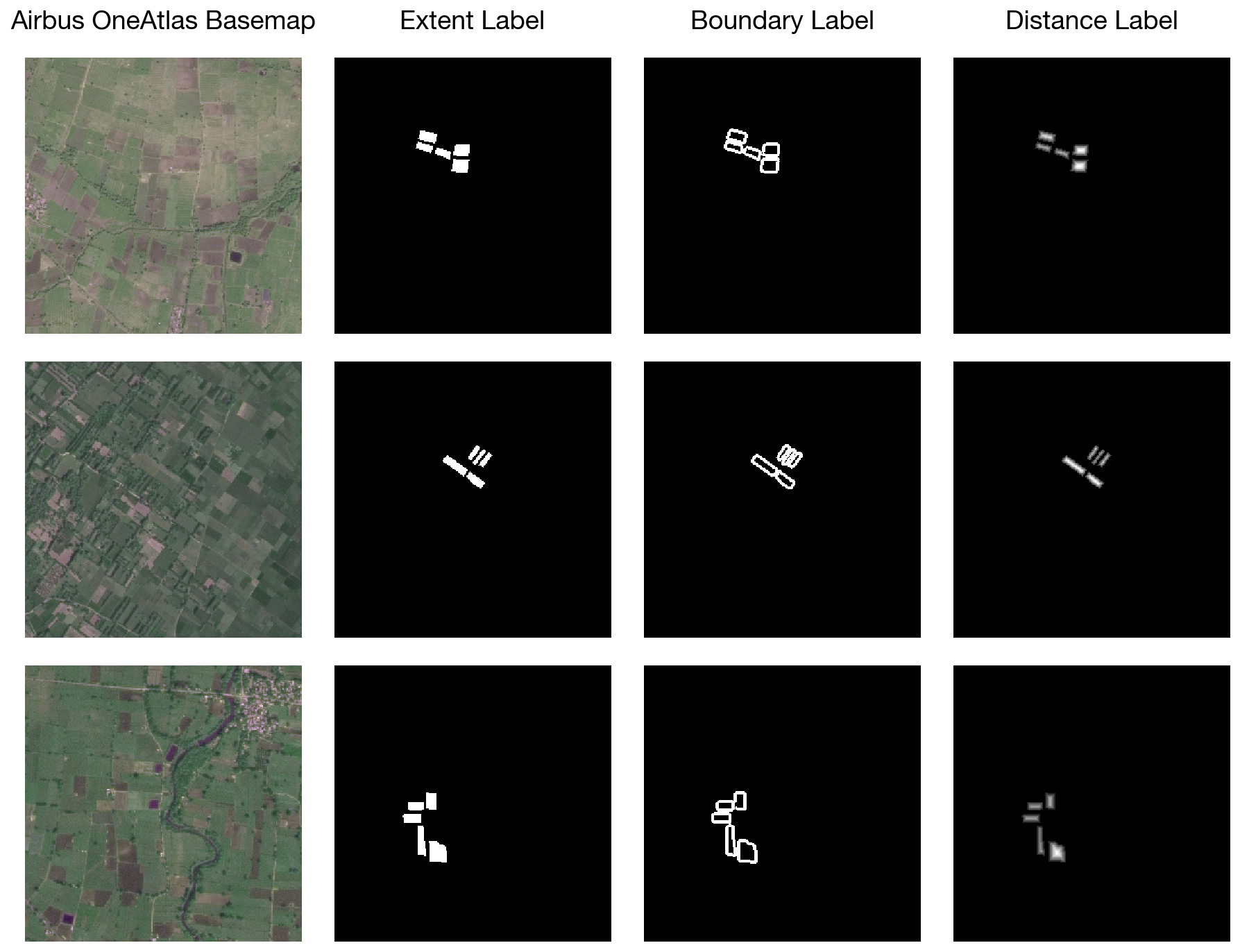}
	\caption{\textbf{Examples of labels used for multi-task learning.} Following the field delineation model design in \citet{waldner2021detect}, we trained all neural networks in this paper to predict three tasks simultaneously: field extent, field boundary, and distance to field boundary.}
	\label{appendix:fig:labels}
\end{figure}
\begin{figure}[h]
	\centering
	\includegraphics[width=0.7\linewidth]{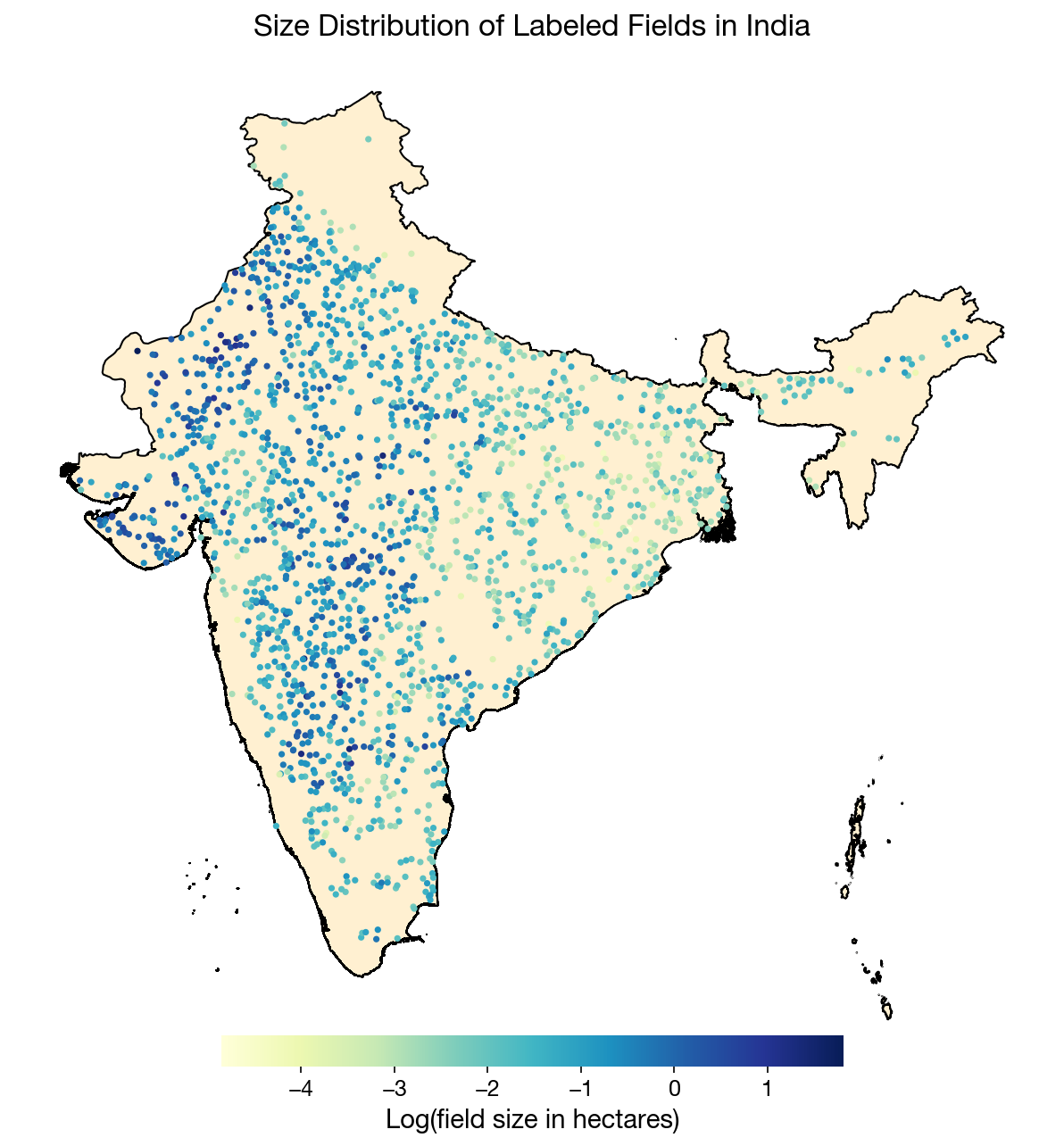}
	\caption{\textbf{India field size distribution from collected labels.} Each point represents a Geo-Wiki location where five field labels were collected. The average field size in each image is plotted on a log scale.}
	\label{appendix:fig:fieldsize}
\end{figure}
\begin{figure}[h]
	\centering
	\includegraphics[width=0.9\linewidth]{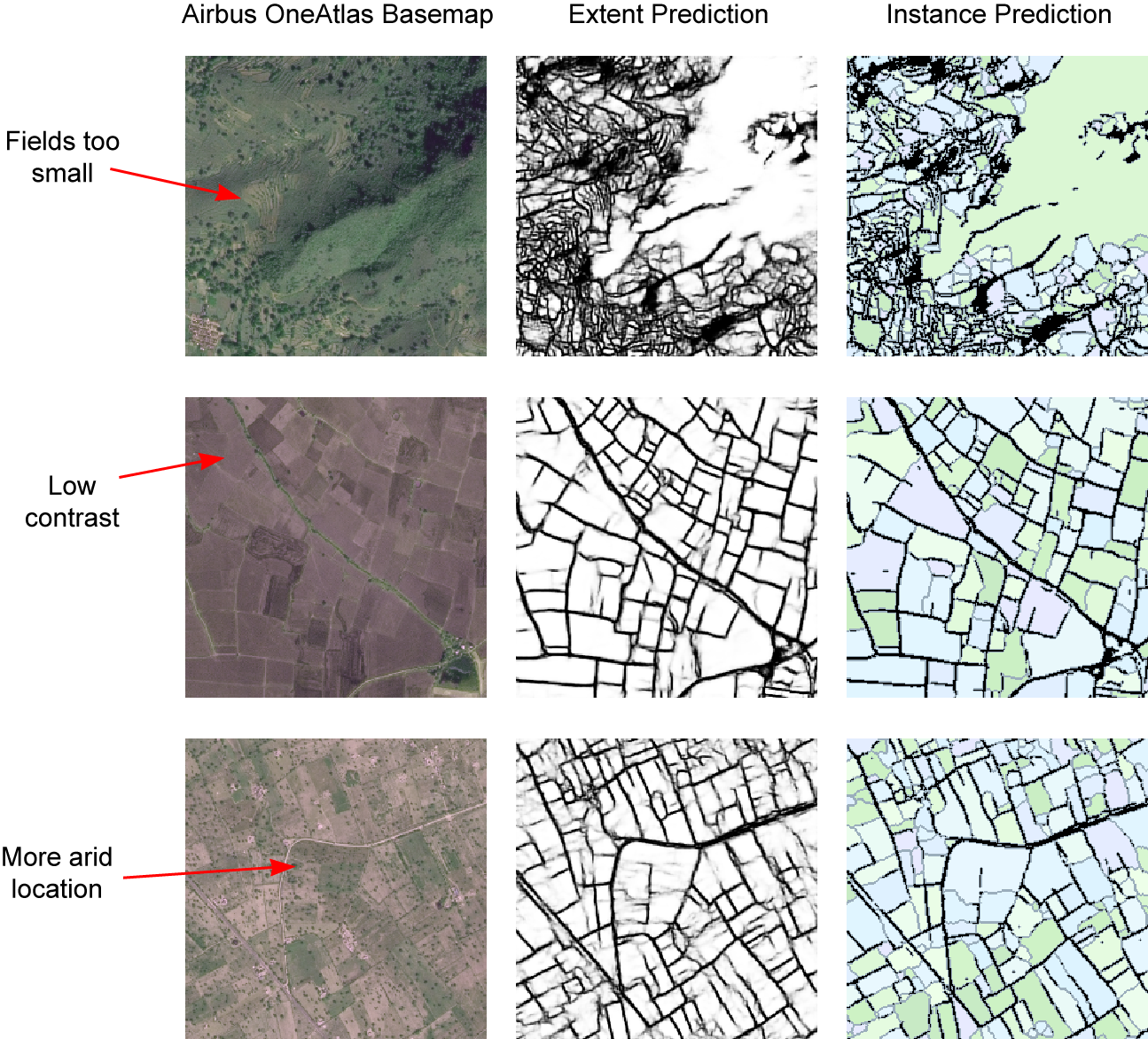}
	\caption{\textbf{Example locations where the model performed poorly.} Commonly seen sources of error include field size being too small, fields having low contrast at the time of image acquisition, and more arid growing conditions. More arid fields are likely harder for the model to delineate because the dataset contains fewer of these samples.}
	\label{appendix:fig:errors}
\end{figure}

 \bibliographystyle{elsarticle-num-names} 
 \bibliography{cas-refs}

\end{document}